\documentclass[10pt,twocolumn,letterpaper]{article}

\usepackage{iccv}              % To produce the CAMERA-READY version
\usepackage[accsupp]{axessibility}  % Improves PDF readability for those with disabilities.
\usepackage{graphicx}
\usepackage{amsmath}
\usepackage{amssymb}
\usepackage{booktabs}
\usepackage{multirow}
\usepackage{color}
\usepackage{pifont}
\usepackage[T1]{fontenc}
\usepackage{dsfont}
\usepackage{rotating} 
\usepackage{marvosym}
\usepackage[ruled,vlined]{algorithm2e}

\DeclareUnicodeCharacter{2113}{\ell}

\definecolor{cvprblue}{rgb}{0.21,0.49,0.74}
\usepackage[pagebackref,breaklinks,colorlinks,allcolors=cvprblue]{hyperref}

\def\paperID{10688} % *** Enter the Paper ID here
\def\confName{ICCV}
\def\confYear{2025}

\title{Noise2Score3D: Tweedie's Approach for Unsupervised Point Cloud Denoising}

\author{
    Xiangbin Wei\textsuperscript{\rm 1} \quad
    Yuanfeng Wang\textsuperscript{\rm 2}\quad
    Ao XU\textsuperscript{\rm 3} \quad
    Lingyu Zhu\textsuperscript{\rm 4} \quad \\
    Dongyong Sun\textsuperscript{\rm 4}  \quad
    Keren Li\textsuperscript{\rm 1 \rm2} \quad 
    Yang Li\textsuperscript{\rm 5} \quad 
    Qi Qin \textsuperscript{\rm 1 \rm2 \rm6}\thanks{Corresponding authors}\footnotemark[1] \\
    \textsuperscript{\scriptsize{\rm 1}}\normalsize{Shenzhen University} \quad
    \textsuperscript{\scriptsize{\rm 2}}\normalsize{Quantum Science Center of Guangdong-Hong Kong-Macao Greater Bay Area} \quad \\
    \textsuperscript{\scriptsize{\rm 3}}\normalsize{Research Institute of Tsinghua University in Shenzhen} \quad 
    \textsuperscript{\scriptsize{\rm 4}}\normalsize{YunJi Intelligent Engineering Co., Ltd.} \quad \\ 
    \textsuperscript{\scriptsize{\rm 5}}\normalsize{Nanjing University} \quad
    \textsuperscript{\scriptsize{\rm 6}}\normalsize{City University of Hong Kong} \quad
    }

\begin{document}
\maketitle

%%%%%%%% ABSTRACT
\begin{abstract}
Building on recent advances in Bayesian statistics and image denoising, we propose Noise2Score3D, a fully unsupervised framework for point cloud denoising. Noise2Score3D learns the score function of the underlying point cloud distribution directly from noisy data, eliminating the need for clean data during training. Using Tweedie's formula, our method performs denoising in a single step, avoiding the iterative processes used in existing unsupervised methods, thus improving both accuracy and efficiency. Additionally, we introduce Total Variation for Point Clouds as a denoising quality metric, which allows for the estimation of unknown noise parameters. Experimental results demonstrate that Noise2Score3D achieves state-of-the-art performance on standard benchmarks among unsupervised learning methods in Chamfer distance and point-to-mesh metrics. Noise2Score3D also demonstrates strong generalization ability beyond training datasets. Our method, by addressing the generalization issue and challenge of the absence of clean data in learning-based methods, paves the way for learning-based point cloud denoising methods in real-world applications.
\end{abstract}

%%%%%%%% BODY TEXT
\section{Introduction}
\label{sec:intro}
%-------------------------------------------------------------------------
\begin{figure*}[ht!]
  \centering
    \includegraphics[width=1.0\textwidth]{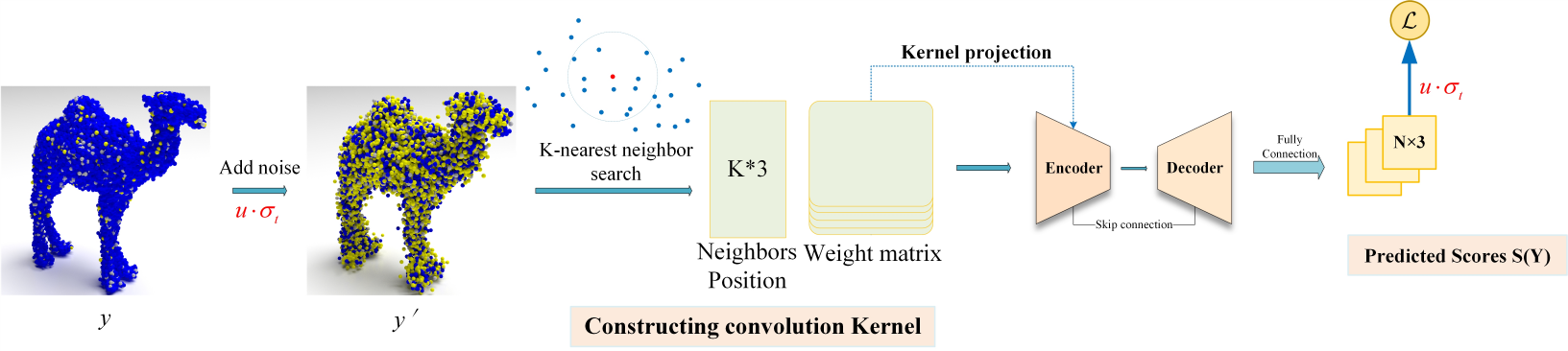}

  \caption{Training workflow of Noise2Score3D with feature extraction and score prediction by an encoder-decoder network with the AR-DAE loss. Denoising is done with Tweedie's formula using estimated scores to restore the position of point clouds (not shown here).}
  \label{fig:kpconv}
\end{figure*}
%\vspace{-0.2cm}
%-------------------------------------------------------------------------
With the proliferation of 3D scanners and depth cameras, the capture and processing of 3D point clouds have become commonplace \cite{quan2024deep}. However, the captured data are often corrupted by noise due to factors like sensor errors and environmental conditions, which can significantly affects downstream tasks.
Removing noise from point clouds—which consist of discrete 3D points sampled irregularly from continuous surfaces—is a long-standing challenge due to their irregular and unstructured nature and the difficulty in distinguishing noise from fine geometric details. Existing work includes traditional optimization-based approaches and deep learning-based approaches. Traditional methods \cite{computingpointset2003,definingpointset2004,wlop2009HH,pointsetsurfaces2001,Robustmoving2005, digne2017bilateral,l1sparse2010,clop2014PR,multi-projection2018duan,Lu2020lowrank, zeng2019GLR} have the advantage of lower computational complexity and less data dependency, but they often rely on predefined geometric priors and cumbersome optimization techniques. As a consequence, it is sometimes challenging to strike a balance between the detail preservation and denoising effectiveness. 

With the continuous emergence of neural network architectures designed for point clouds \cite{qi2017pointnet,qi2017pointnet2,KPconv,wang2019dynamic}, deep learning-based denoising methods have recently shown promising performance \cite{NPD2019,rakotosaona2020PCN,hermosilla2019TotalDenoising,luo2020DMR,luo_score-based_2021,10173632}. 
In supervised settings, training denoising models requires pairs of noisy and clean data. However, in many real-world scenarios, obtaining clean data is difficult or impractical. Therefore, unsupervised denoising has become a compelling research area. Meanwhile, various unsupervised image denoising methods have been proposed \cite{2018Noise2NoiseLI,2018Noise2VoidL,2018Noise2VoidL,2019Noise2SelfBD,kim_noise2score_2021} \textit{etc.} 
 %Among those, Noise2Score \cite{kim_noise2score_2021} introduces a novel method for self-supervised image denoising for data with arbitrary exponential family noise. The Tweedie's formula \cite{efron2011tweedie} provides an explicit way to compute posterior estimation from noisy measurements with exponential family noise given the score function (\textit{i.e.}, the gradient of the log-likelihood).
These methods are particularly important in practical applications where obtaining clean images or multiple noisy realizations of the same image is difficult or impossible.

Inspired by this, we propose Noise2Score3D, an unsupervised point cloud denoising method that, using only noisy point clouds as input, predicts the score function and applies the Tweedie's formula to estimate the clean point positions based on the predicted score. Our method achieves state-of-the-art performance among unsupervised methods and offers improved generalization ability across different datasets, as well as noise levels without re-training. Noise2Score3D also achieves high efficiency, as denoising is performed in one step with known noise parameters. Our source-code is publicly available at 
\href{https://anonymous.4open.science/r/denoise-FE82}{here}.

The main contributions of this paper are as follows:
\begin{itemize}
\item We propose a method for estimating the score function of point clouds from noisy inputs based on the amortized residual denoising autoencoder with the KPConv network architecture \cite{KPconv}. 
\item Combining score estimation with Tweedie's formula, we propose an unsupervised learning framework for point cloud denoising. Experiments demonstrate that Noise2Score3D achieves state-of-the-art results on several benchmark datasets with varying noise levels in terms of both quantitative metrics and visual quality. 
\item We introduce total variation for point cloud, a metric for assessing the quality of denoised point clouds, enabling estimation of unknown noise parameters, thereby making our method broadly applicable to real-world data.
\end{itemize}

\section{Related Work}
\label{sec:related work}
% In this section, we review the existing literature on point cloud denoising and unsupervised image denoising.
%-------------------------------------------------------------------------
\subsection{Point cloud denoising}

    \subsubsection{Traditional methods}
Traditional approaches to point cloud denoising can be categorized into statistical methods \cite{computingpointset2003,definingpointset2004,wlop2009HH}, filtering-based methods \cite{Robustmoving2005, digne2017bilateral}, and optimization-based methods \cite{l1sparse2010,clop2014PR,multi-projection2018duan,Lu2020lowrank, zeng2019GLR}. Statistical methods operate under the assumption that noise is statistically different from the signal and use presumed distribution models to identify and remove outliers. Filtering methods, such as \cite{digne2017bilateral}, adapt the well-developed bilateral filtering method in image analysis to point clouds.  Optimization-based methods formulate denoising as an energy minimization problem, where certain regularization terms constrain the solution to ensure a certain smoothness criterion or adherence to prior knowledge. For instance, Lu \etal.\cite{Lu2020lowrank} developed a low-rank approximation-based method on estimated surfaces normals for point cloud filtering. Zeng \etal. \cite{zeng2019GLR} exploited the self-similarity of the surface patches and proposed a patch-based graph Laplacian regularizer to denoise collaboratively, which shows strong denoising performance and structural detail preservation. 

%-------------------------------------------------------------------------
    \subsubsection{Deep learning methods}
In recent years, several deep learning-based methods  have been proposed for point cloud denoising. Duan \etal. \cite{NPD2019}  proposed the first learning-based point cloud denoising method that directly processes noisy data without requiring noise characteristics or neighboring point definitions.  PointCleanNet \cite{rakotosaona2020PCN} first removes outlier points and then combines them with residual connectivity to predict the inverse displacement \cite{Guerrero2017PCPNetLL}, and iteratively shifts noisy points to remove noise. Luo \etal also proposed DMRDenoise \cite{luo2020DMR}, which filter points by first downsampling the noisy inputs and reconstructing the local subsurface to perform point upsampling. However, the resampling procedure is difficult to maintain a good local shape. ScoreDenoise \cite{luo_score-based_2021} is proposed to tackle the aforementioned issues by updating the point position in implicit gradient fields learned by neural networks. During inference, they follow an iterative procedure with a decaying step size, which stabilizes point movement and prevents over-correction, allowing points to converge gradually toward the underlying geometry. IterativePFN \cite{de_Silva_Edirimuni_2023_CVPR} uses a novel loss function that utilizes an adaptive ground truth target at each iteration to capture the relationship between intermediate filtering results during training. Zheng \etal proposed an end-to-end network for joint normal filtering and point cloud denoising \cite{10173632}, which achieves good performance in both point cloud denoising and normal estimation. 

While achieving impressive results, supervised methods are limited by the availability and quality of the training data, as they typically require paired noisy and clean point clouds to train the model. Instead, unsupervised learning-based approaches leverage the inherent structure or distribution of the point cloud to guide the denoising process and show promise in scenarios where clean data is absent or hard to obtain. Total Denoising (TotalDn) \cite{hermosilla2019TotalDenoising} is the first unsupervised learning approach for point cloud denoising, relying solely on noisy data. TotalDn approximates the underlying surfaces by regressing points from the distribution of unstructured total noise, utilizing a spatial prior term to refine the geometry. An unsupervised version of DMRDenoise \cite{luo2020DMR} utilizes a loss function that identifies local neighborhoods using a probabilistic Gaussian mask on the k-nearest neighbors, which selectively retains points likely to represent the underlying surface.
% By leveraging an Earth Mover's Distance (EMD) assignment, it achieves a one-to-one correspondence between input and predicted points, aligning them to reduce noise within local neighborhoods. This approach enhances robustness to noise and adapts well to varied surface geometries. However, the probabilistic masking and EMD calculation add computational complexity, which can slow down inference in dense or noisy point clouds. 

ScoreDenoise \cite{luo_score-based_2021} also introduced an unsupervised version  (Score-U) that employs an ensemble score function with an adaptive neighborhood-covering loss for model training. However, the defined score in \cite{luo_score-based_2021} is still based on displacements instead of likelihood, and thus unable to utilize tools in Bayesian statistics. The denoising process, which is done iteratively with the estimated gradients, can be computationally expensive. Overall, unsupervised learning-based point cloud denoising methods are still in the early stages. Several key issues remain to be addressed, including model generalization (to different datasets and noise levels beyond training data) and inference efficiency \cite{zhou2022point,quan2024deep}. 
% Most recently, Noise4Denoise \cite{noise4Wang2024} method is proposed that use an additional doubly-noisy point cloud to learn noise displacement by comparing the two noise levels. This approach effectively leverages synthetic noise for training, allowing the model to estimate residuals without relying on clean data.
% However, in practical applications, noise parameters are often unknown and variable, making it challenging to replicate the exact conditions assumed during training. This reliance on predefined noise characteristics can limit the model's applicability to real-world scenarios where noise distributions may differ significantly from synthetic settings. 

%-------------------------------------------------------------------------
\subsection{Score matching and estimation}
Given an underlying data distribution, score matching aims to match the model-predicted gradient and the log-likelihood gradient of data \cite{hyvarinen2005estimation}.  It has wide applications in generative models \cite{song2019generative,yang2023diffusion}, and has recently been applied to 3D shape generation \cite{cai2020learning}. Score function estimation has been a significant research topic in Bayesian statistics and machine learning \cite{hyvarinen2005estimation, vincent2011connection, alain2014regularized}. In particular, Alain and Bengio \cite{alain2014regularized} showed that minimizing the objective function of a denoising autoencoder (DAE) provides an explicit way to approximate the score function. Later, an amortized residual denoising autoencoder (AR-DAE) \cite{lim2020ar-dae} is proposed to enable a more reliable score estimation.

%-------------------------------------------------------------------------
\subsection{Unsupervised image denoising}
Recently, unsupervised image denoising has made significant progress. Non-Bayesian methods include PURE \cite{luisier2010image}, SURE \cite{SURE2018} \textit{etc.}, which are based on various unbiased risk estimators under certain noise distributions. Other methods explore self-similarity in natural images \cite{xu2015patch, doi:10.1137/23M1614456} or exploits the statistical properties of noise to achieve a denoising effect \cite{gravel2004method}.  

Noise2Noise \cite{2018Noise2NoiseLI} is a pioneering method that does not require clean data, but it requires multiple noisy versions of the same image for training. To address this limitation, methods such as Noise2Void \cite{2018Noise2VoidL}, Noise2Self \cite{2019Noise2SelfBD}, \textit{etc.}, have been developed, which use only a single noisy image. These methods are trained by minimizing variants of empirical risk.  Neighbor2Neighbor \cite{huang2021neighbor2neighbor} proposed a two-step method with a random neighbor sub-sampler that generates training  pairs and a denoising network. Kim \etal proposed Noise2Score \cite{kim_noise2score_2021}, a novel Bayesian framework for self-supervised image denoising without clean data. The core of Noise2Score is the usage of Tweedie's formula, which provides an explicit representation of the denoised image through a score function. By reframing image denoising as the problem of score function estimation, Noise2Score offers a new perspective that provides important theoretical implications and flexibility in algorithmic implementation.

% Combined with score function estimation, Noise2Score can be applied to image denoising with any exponential family noise. Kim \etal also proposed the Noise Distribution Adaptive Self-Supervised Image Denoising method \cite{kim_noise_2022}, which further extends the application of Noise2Score by combining the Tweedie distribution with score matching. This enables adaptive handling of various noise distributions and dynamically adjusts the denoising process by estimating noise parameters.

% These development of unsupervised image denoising method motivate us to explore similar ideas in 3D point cloud denoising.

% \begin{figure}
% \begin{center}
%     \includegraphics[width=0.5\textwidth]{figures/diff_denoise.pdf}
% \end{center}
% \vspace{-1.0em}
% \caption{Illustration of different classes of learning based methods.}
% \label{fig:diff_method}
% \end{figure}
%-------------------------------------------------------------------------
% \begin{figure}[t]
%   \centering
%   \includegraphics[width=1.0\linewidth]{figures/diff_denoise.pdf}
%   \caption{Illustration of different classes of learning-based methods. Hollow circles represent the initial positions of data points, while filled circles indicate adjusted positions.}
%   \label{fig:diff_method}
% \end{figure}

\section{Method}
\label{sec:method}
Our proposed framework, e.g. Noise2Score3D, adopts a Bayesian approach to unsupervised point clouds denoising. Specifically, Noise2Score3D consists of two steps: 1) estimating the score function from the noisy point cloud, and 2) restoring the clean point cloud using the estimated score and noise parameters.
This decoupling between model estimations and denoising (displacements calculation) allows Noise2Score3D to have unique advantages over other unsupervised learning methods, the most important of which is that the same loss function or pre-training weight can be used regardless of the noise model and parameters. By utilizing Tweedie's formula from Bayesian statistics, the denoising process can be done in one step, given the noise parameters. These features distinguish our methods from other unsupervised denoising methods, such as TotalDn \cite{hermosilla2019TotalDenoising} and ScoreDenoise \cite{luo_score-based_2021}. As a result, our approach is simple to train and use, and does not expect the user to provide parameters to characterize the surface or noise model. In \cref{sec:method:theory}, we first introduce the basic theory. 
In \cref{sec:method:scoreest}, we elaborate on each proposed step, including the score estimation framework with the corresponding network structure  and loss function. Finally,  in \cref{sec:method:denosing}, we describe the denoising process, emphasizing on the case of unknown noise parameters.

%-------------------------------------------------------------------------
\subsection{Theory}
\label{sec:method:theory}
\textbf{Problem definition}\quad Assume that the noisy point cloud data $\mathbf{y}$ follows some exponential family distribution whose density has the following general form:
\begin{eqnarray}
p(\mathbf{y}\mid\eta) = b(\mathbf{y}) \exp\{\eta^T T(\mathbf{y})-\phi(\eta) \}.
\end{eqnarray}
where $\eta$ is the canonical parameter related to the clean data.  $T(\mathbf{y})$ and $\eta$ have the same dimensions, and $\phi(\eta)$ and $b(\mathbf{y})$ are both scalar functions, with $b(\mathbf{y})$ being the density up to a scale factor when $\eta = 0$.

Using the Bayesian formula, the posterior distribution $\eta$ has the following form:
\begin{eqnarray}
p(\eta\mid\mathbf{y})=p(\eta)e^{-\phi(\eta)} \exp(\eta^T T(\mathbf{y}) +\log \frac{b(\mathbf{y})}{p(\mathbf{y})}).
\end{eqnarray}

The problem is to estimate the posterior mean of $\eta$, e.g. $E_{p(\eta \mid \mathbf{y})}(\eta)$ or the mode of the distribution.
% \begin{eqnarray}
% y = x + w, \quad w \sim N(0, \sigma^2I).
% \end{eqnarray}
% where $w$ represents independent and identically distributed Gaussian noise with zero mean and variance $\sigma^2$.
%-------------------------------------------------------------------------
\vspace{0.2cm}

\noindent \textbf{Tweedie's formula for denoising}\quad Tweddie's formula \cite{efron2011tweedie} provides an explicit way for calculating the posterior expectation of canonical parameters or variables corrupted with exponential family noises via the gradient of the log-likelihood function, i.e., the score function of the noisy data distribution. In our approach, we extend this concept to the 3D point cloud denoising task.

Tweedie’s formula states that the posterior estimate $\hat{\eta}$ is given by the following equality:
\begin{eqnarray}
\nabla_y \log p(\mathbf{y}) = \nabla_y \log b(\mathbf{y}) + T'(\mathbf{y})^T \hat{\eta}.
\end{eqnarray}
where $T'(\mathbf{y})=\nabla_{\mathbf{y}} T(\mathbf{y})$. For the special case of additive Gaussian noise, we have $T(\mathbf{y}) = \mathbf{y}$, $\eta = \mathbf{x}/\sigma^2$  and $b(\mathbf{y}) = \frac{1}{\sqrt{2\pi}\sigma} e^{-\frac{y^2}{2\sigma^2}}$. The posterior expectation $\mathbf{x}$ is given by \cite{noise2score2021}:
\begin{eqnarray}
E_{p(\mathbf{x}\mid\mathbf{y})}(\mathbf{x}) = \mathbf{y} + \sigma^2 \nabla_\mathbf{y} \log p(\mathbf{y}).
\end{eqnarray}
where $\nabla_\mathbf{y} \log p(\mathbf{y})$ is the score function of $\mathbf{y}$ the data distribution.
% This give the optimal posterior estimation of $X$ in the sense of Minimum Mean Squared Error (MMSE)\cite{Serfling1980ApproximationTO}.

The posterior form for other types of exponential family noise can be found in \cite{kim_noise2score_2021}. More generally, Tweedie’s formula has been extended to more complex noise distribution \cite{scoreXie2024} and provides an unified approach to denoising data with various noise types.
% As noted in \cite{noise2score2021}, efficient denoising can be achieved by utilizing sufficient statistics of the noise distribution even if the noise parameters are unknown. It is note-worthy that once trained, there is no need to retrain the neural network for different noise level, making our denoising algorithm generalizing better compared to traditional methods.

%-------------------------------------------------------------------------
\subsection{Score estimation}
\label{sec:method:scoreest}
\textbf{Network for score estimation}\quad We build upon the Kernel Point Convolution (KPConv) architecture introduced by Thomas \etal \cite{KPconv} for score estimation. 
KPConv network is a point cloud feature encoder which accurately captures local geometric structures and adapts to varying point densities. We modified it to accept only the three-dimensional coordinates of the noisy point cloud $y$ as input features. 
The convolution weights of KPConv are determined by Euclidean distances to kernel points, and the number of kernel points is set to 15 in our implementation. 
% The positions of these kernel points are formulated as an optimization problem for best coverage in a spherical space. 
Note that the radius neighborhood (e.g., increasing from 0.1 to 1.6 across stages) is used to maintain a consistent receptive field, while grid subsampling is applied in each layer to achieve high robustness under varying densities of point clouds. Our network comprises five encoder stages with a total of 14 blocks: an initial KPConv block and a residual block in the first stage, followed by four stages, each with three residual blocks outputting 256, 512, 1024, and 2048 channels, respectively. The decoder includes four blocks, reducing the feature dimension from 3072 to 128, and a final fully connected layer outputs three-dimensional vectors (estimated score values) for each point. The network architecture is illustrated in \cref{fig:kpconv} with total parameter-number of 24.3M. 
The output score values are interpreted as the values statistically closest to the gradient of the log-probability density function $S(y)$ at each point.
The estimated score function is used for reconstructing the clean point cloud during the denoising stage, as detailed in the later part of \cref{sec:method:theory}.

\noindent\textbf{Loss function} \quad
The loss function is crucial for training the network to estimate the score function.
In the unsupervised setting, we assume that clean point cloud $x$ is not accessible during training. Therefore, any loss function that requires knowledge of $x$ is not feasible. Instead, we need a loss function that relies solely on the noisy observations $y$.

To achieve this goal, we employ an unsupervised learning method inspired by the Denoising Score Matching proposed by Lim \cite{vincent2011connection}, and the Amortized Residual Denoising Autoencoder (AR-DAE) \cite{lim2020ar-dae}. Specifically, the loss function is defined as follows:
\begin{eqnarray} 
\mathcal{L}_{\text{AR-DAE}} = \frac{1}{N} \sum_{i=1}^{N} \left\| \sigma_t \cdot S(y'_i) + u \right\|^2.
\end{eqnarray}
where
$N$ is the number of points in the point cloud,
$y'_i$ is a perturbed version of a point $y_i$ in the noisy point cloud $y$, generated by adding noise: $y'_i = y_i + u \cdot \sigma_t$, where $u \sim \mathcal{N}(0, I)$ represents random Gaussian noise. $\sigma_t$ is the noise standard deviation during training. $S(y'_i)$ is the estimated score at the perturbed point $y'_i$.
$u$ is an additional random noise vector sampled from $\mathcal{N}(0, I)$.

The loss function aims to minimize the difference between the scaled estimated score $\sigma_t^2 \cdot S(y'_i)$ and the additional noise $u$. By doing so, the network learns to approximate the negative residual noise scaled by the noise level, which corresponds to the score function of the noisy data distribution.

%-------------------------------------------------------------------------
\begin{algorithm}[!h]
\caption{Denoising with unknown noise parameters}
\KwIn{Noisy point cloud $y \in \mathbb{R}^{n \times 3}$, trained model $\mathbf{F}$ for score prediction, sigma range $\Sigma = \{\sigma_1, \sigma_2, \dots, \sigma_k\}$}
\KwOut{Denoised point cloud $\hat{x}$}

Predict scores $S(y) = \mathbf{F}(y)$\;
Initialize $\sigma^* \gets 0$, $TV_{pc}^* \gets \infty$, $\hat{x} \gets y$\;

\For{each $\sigma \in \Sigma$}{
    $x(\sigma) = y + \sigma^2 S(y)$\;
    Calculate $TV_{pc}(\sigma)$ for $x(\sigma)$\;
    
    \If{$TV_{pc}(\sigma) < TV_{pc}^*$}{
        $TV_{pc}^* \gets TV_{pc}(\sigma)$\;
        $\sigma^* \gets \sigma$\;
        $\hat{x} \gets x(\sigma)$\;
    }
}
\Return $\hat{x}$\;
\label{algorithm:1}
\end{algorithm}

%-------------------------------------------------------------------------

\subsection{Denoising with unknown noise parameters}
\label{sec:method:denosing}

Our method depends on a known or good estimate of the noise parameters. For real-world scenarios, the noise parameter/level of point clouds is often unknown. To deal with this case, we proposed a novel quality metric for point cloud denoising, namely Total Variation for Point Cloud. Based on that, we use an in-loop denoising procedure to estimate the unknown noise parameters. 

We extend the total variation concept in image denoising \cite{RUDIN1992} to point cloud processing and define Total Variation for Point Cloud ($TV_{PC}$) as: 
\begin{equation} 
\label{eq:TV_{pc}_main}
TV_{PC} = \sum_{i=1}^N \sum_{j \in \text{neighbors}(i)} w_{i,j} \cdot \sqrt{\| \mathbf{p}_i - \mathbf{p}_j \|^2 + \epsilon^2}.
\end{equation}
where \( w_{i,j} \) represents the weight between point \( \mathbf{p}_i \) and its neighbor \( \mathbf{p}_j \), $\epsilon$ is a small positive constant. Our definition of $TV_{PC}$  measures the geometric difference between a point $\mathbf{p}_i$ and its $k$ nearest neighboring points, which favors smooth surfaces. The detailed description and validation of $TV_{PC}$ is provided in the Supplementary Material. 
We estimate the noise parameter \( \sigma^* \) by minimizing the $TV_{PC}$ resulting point cloud:
\begin{equation}\label{eq:sigma}
\sigma^* = \arg\min\limits_\sigma TV_{PC}\left(\hat{x}(\sigma)\right).
\end{equation}
where $\hat x(\sigma) = y + \sigma^2 S(y)$.
This allows us to automatically find the best noise parameter from the data, ensuring that the processed point cloud achieves optimal smoothness while minimizing geometric distortion. The pseudo code for denoising using $TV_{PC}$ is shown in \cref{algorithm:1}.

%-------------------------------------------------------------------------
\begin{table*}[htbp!]
\caption{Comparison of the denoising performance on ModelNet-40 with Gaussian noise by different algorithms. The values for CD and P2M are multiplied by $10^4$. The best (second-best) results are shown in \textbf{bold} (\underline{underlined}).}
\vspace{-0.5cm}
\begin{center}
\resizebox{\textwidth}{!}{
\begin{tabular}{cc|cccccc|cccccc}
\hline
\multicolumn{2}{c|}{Number of points}         & \multicolumn{6}{c|}{Gaussian 10k} & \multicolumn{6}{c}{Gaussian 50k}  \\
\multicolumn{2}{c|}{Noise level}     & \multicolumn{2}{c}{1\%} & \multicolumn{2}{c}{2\%} & \multicolumn{2}{c|}{3\%} & \multicolumn{2}{c}{1\%} & \multicolumn{2}{c}{2\%} & \multicolumn{2}{c}{3\%} \\ \cline{3-14} 
\multicolumn{1}{c|}{Dataset}   & Model  & CD & P2M & CD & P2M & CD & P2M & CD & P2M & CD & P2M & CD & P2M   \\ \hline
\multicolumn{1}{c|}{\multirow{7}{*}{Modelnet-40}} 
& Bilateral\cite{digne2017bilateral} & 5.865      & \underline{3.016}      & 15.121      & 9.441     & 31.034      & 21.974      & 3.711      & 3.192      & 13.466      & 11.308      & 30.194      & 25.669      \\
\multicolumn{1}{c|}{}                    & Lowrank\cite{Lu2020lowrank}   & 5.698      & 3.069      &\underline{8.882}      &\underline{4.678}      & 14.846      & 8.629          & \bf1.796      & \bf1.702      & 6.448      & \underline{5.121}      & 17.670       & \underline{14.332} \\        
    \multicolumn{1}{c|}{}                    & GLR\cite{zeng2019GLR}   & 6.592      & 3.700      &\bf8.365      & \bf4.582      & \underline{12.890}      & \bf7.877          & \underline{1.860}      & \underline{1.852}      &\underline{6.147}      & 5.169      & \underline{17.462}       & 14.629        \\ \cline{2-14} 
\multicolumn{1}{c|}{}  & TotalDn\cite{hermosilla2019TotalDenoising} &8.079 &4.778  &18.031 &12.277  &29.617 &21.673     & 5.044& 4.442  & 13.130  & 11.165      & 22.627      & 19.334            \\

\multicolumn{1}{c|}{}  & DMR-U\cite{luo2020DMR} &8.210 &5.044  &12.770 &8.201  &22.086 &15.602  & 3.250 & 2.946  & 10.430 & 8.975  & 23.596 & 20.426    \\
\multicolumn{1}{c|}{}  & Score-U\cite{luo_score-based_2021} &\underline{5.514} &\bf2.975  &11.072 &6.412  &18.239 &11.335  & 2.696 & 2.317   & 10.153  &  8.269   & 26.169  & 21.664  \\ \cline{2-14} 

\multicolumn{1}{c|}{} & \bf Ours  &\bf5.283 &3.212      &8.624 &5.332   &\bf12.791 &\underline{8.568}      & 1.881 &1.988  &\bf4.393  &\bf4.002   &\bf8.917  &\bf8.181  \\ \hline
\end{tabular}}
\end{center}
\label{table:quantitativemd}
\end{table*}
%-------------------------------------------------------------------------

With the proposed $TV_{PC}$ as point cloud quality assessment metrics, we can estimate the unknown parameters for real-world datasets, resulting in a fully unsupervised framework for point cloud denoising applications.

\section{Experiments}
\label{sec:experiment}

% \usepackage{multirow}
\begin{table*}[tb]
\caption{Comparison of denoising performance of different algorithms on PU-Net. The values for CD and P2M are multiplied by $10^4$. The best (second-best) results are shown in \textbf{bold} (\underline{underlined}).}
\vspace{-0.5cm}
\begin{center}
\resizebox{\textwidth}{!}{
\begin{tabular}{cc|cccccc|cccccc}
\hline
\multicolumn{2}{c|}{Number of points}                     & \multicolumn{6}{c|}{10k}                                                     & \multicolumn{6}{c}{50k}                                                     \\
\multicolumn{2}{c|}{Noise level}                 & \multicolumn{2}{c}{1\%} & \multicolumn{2}{c}{2\%} & \multicolumn{2}{c|}{3\%} & \multicolumn{2}{c}{1\%} & \multicolumn{2}{c}{2\%} & \multicolumn{2}{c}{3\%} \\ \cline{3-14} 
\multicolumn{1}{c|}{Dataset}    & Model & CD & P2M   & CD  & P2M     & CD  & P2M   & CD & P2M   & CD& P2M        & CD         & P2M        \\ \hline
\multicolumn{1}{c|}{\multirow{7}{*}{PU-Net}}    & Bilateral\cite{digne2017bilateral} & 3.646      & 1.342      & 5.007      & 2.018     & 6.998      & 3.557      & 0.877      & \underline{0.234}      & 2.376      & 1.389      & 6.304      & 4.730      \\
\multicolumn{1}{c|}{}                    & Lowrank\cite{Lu2020lowrank}   & 3.820      & 1.575      &4.878      & 2.080      & 5.873      & \underline{2.778}          &\underline{0.778}      & \bf0.176      & \underline{1.303}      & \bf0.506      & 2.551       &\bf1.431 \\  
\multicolumn{1}{c|}{}                    & GLR\cite{zeng2019GLR}   & 3.863      & 1.967      & \underline{4.562}      & 2.194      & \bf5.322       & \bf2.677     & \bf0.767      & 0.251      & \bf1.284      & \underline{0.611}      & \underline{2.446}      & 1.556      \\ \cline{2-14} 
\multicolumn{1}{c|}{}                   & TotalDn\cite{hermosilla2019TotalDenoising} & 3.390      & \bf0.826      & 7.251      & 3.485      & 13.385      & 8.740      & 1.024      & 0.314      & 2.722      & 1.567      & 7.474      & 5.729      \\
\multicolumn{1}{c|}{}                    & DMR-U\cite{luo2020DMR}   & 5.313      & 2.522      & 6.455      & 3.317      & 8.134       & 4.647      & 1.226      & 0.521      & 2.138      & 1.251      & 2.496      & \underline{1.520}      \\
\multicolumn{1}{c|}{}                    & Score-U\cite{luo_score-based_2021} & \underline{3.107}      & \underline{0.888}      & 4.675      & \underline{1.829}      & 7.225       & 3.762      & 0.918      & 0.265      & 2.439      & 1.411      & 5.303      & 3.841      \\ \cline{2-14} 
    \multicolumn{1}{c|}{}    & \bf Ours  & \bf2.848   & 1.106         & \bf4.190         & \bf1.818           & \underline{5.583}     & 2.947           & 0.833           & 0.461           & 1.532           & 0.970          & \bf2.434           & 1.704     \\ \hline
\end{tabular}}
\end{center}
\label{table:quantitativepu}
\end{table*}
\vspace{-0.3cm}
%-------------------------------------------------------------------------

%-------------------------------------------------------------------------
\begin{table}
\caption{Comparison of different denoising algorithms on ModelNet-40 with simulated LiDAR scanner noise.  The values of CD are multiplied by $10^4$. The best (second-best) results are shown in \textbf{bold} (\underline{underlined}).}
\vspace{-0.5cm}
    \begin{center}
    \vspace{-0.05in}
\begin{tabular}{cc|ccc}
\hline
\multicolumn{2}{c|}{Type of noise}         &  \multicolumn{3}{c}{Simulated LiDAR}  \\
\multicolumn{2}{c|}{Noise level} & \multicolumn{1}{c}{0.5\%} & \multicolumn{1}{c}{1.0\%} & \multicolumn{1}{c}{1.5\%} \\ \hline
\multicolumn{1}{c|}{Dataset}   & Model  & CD  & CD  & CD   \\ \hline
\multicolumn{1}{c|}{\multirow{7}{*}{Modelnet-40}}
& Bilateral \cite{digne2017bilateral}  & 1.527 & 1.686  & 2.110     \\
\multicolumn{1}{c|}{}& Lowrank \cite{Lu2020lowrank}  & 1.230 & 1.229  & 1.355    \\
\multicolumn{1}{c|}{}& GLR \cite{zeng2019GLR}  &\bf1.024 & \bf1.074  & \bf1.249     \\\cline{2-5} 
\multicolumn{1}{c|}{}   & TotalDn\cite{hermosilla2019TotalDenoising}   & 1.706     & 2.071  & 2.734        \\
\multicolumn{1}{c|}{}  & DMR-U\cite{luo2020DMR}  & 2.248  & 2.278  & 2.467   \\
\multicolumn{1}{c|}{}  & Score-U\cite{luo_score-based_2021} & 1.469   & 1.426    &1.652    \\ \cline{2-5} 
\multicolumn{1}{c|}{}  & \bf Ours  &  \underline{1.114}   & \underline{1.165}   & \underline{1.327}  \\ \hline
\end{tabular}
\end{center}
\label{table:quantitativemd2}
\end{table}

%-------------------------------------------------------------------------
\begin{figure*}
\begin{center}
    \includegraphics[width=1.0\textwidth]{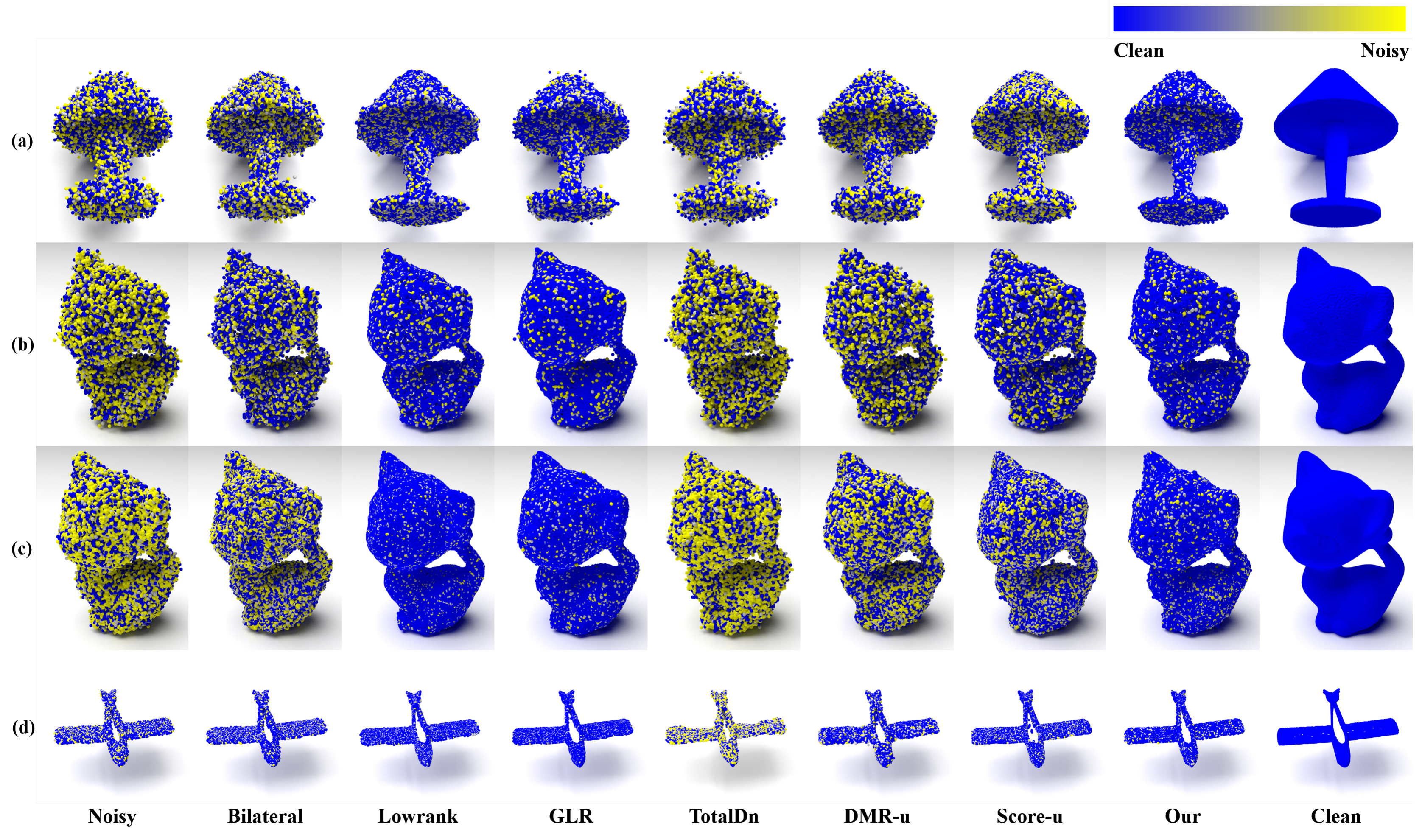}
\end{center}
\vspace{-0.5cm}
\caption{Visualization results by different algorithms on synthetic datasets: (a) ModelNet-40 with 50k points; (b) PU-Net with 10k points; (c) PU-Net with 50k points; (d) ModelNet-40 with simulated LiDAR noise. Note that shown results with the noise scale set to 2\% of the bounding sphere’s radius for the Gaussian noise and 1.5\% for the simulated LiDAR noise. Points with smaller error are colored more blue, and otherwise colored yellow.}
\label{fig:visualization}
\end{figure*}

%-------------------------------------------------------------------------

\begin{figure*}
\begin{center}
    \includegraphics[width=1.0\textwidth]{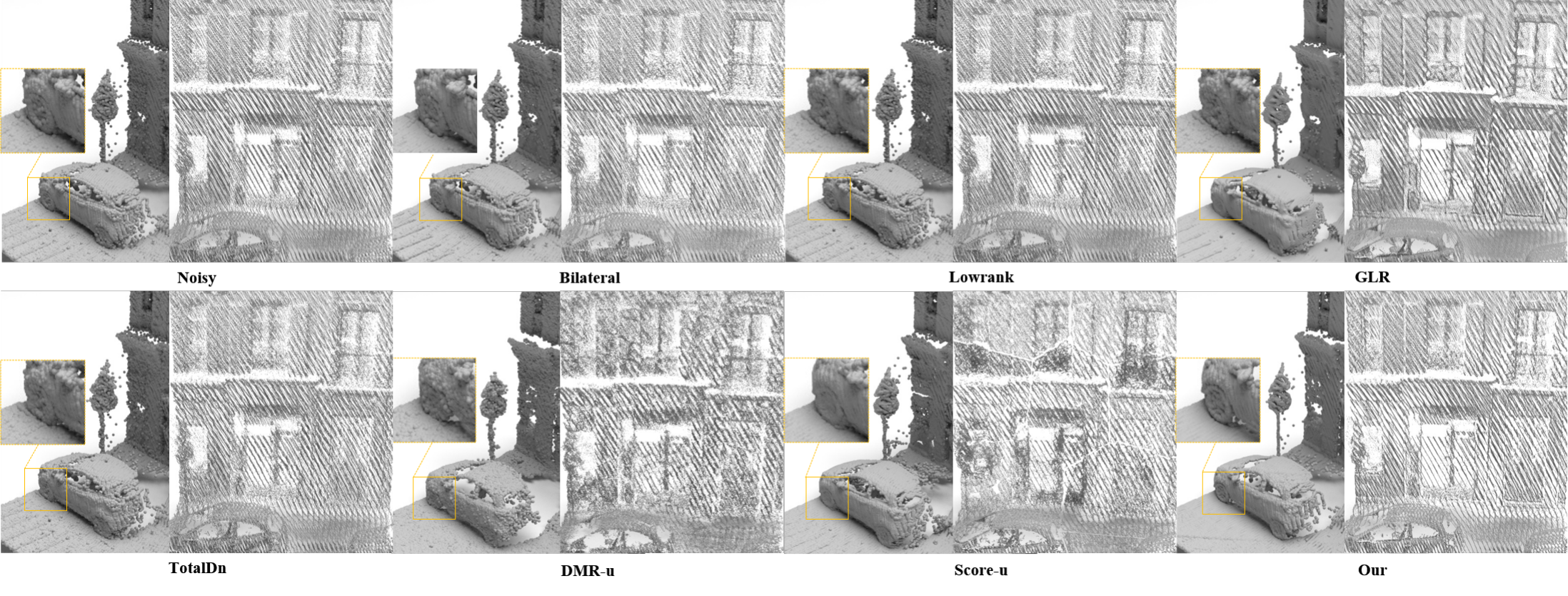}
\end{center}
\vspace{-0.5cm}
\caption{Visualization results on the real-world dataset \textit{Paris-rue-Madame} \cite{serna2014paris}. }
\label{fig:rue32}
\end{figure*}

%-------------------------------------------------------------------------

% \begin{figure*}
% \begin{center}
%     \includegraphics[width=1.0\textwidth]{figures/rue3.pdf}
% \end{center}
% \vspace{-0.05in}
% \caption{Visualization of our denoiser  vs. other algorithms on the real-world dataset \textit{Paris-rue-Madame} \cite{serna2014paris}}
% \label{fig:rue3}
% \end{figure*}

%-------------------------------------------------------------------------

%-------------------------------------------------------------------------
\subsection{Setup}
\label{sec:experiment:setup}

\textbf{Datasets}\quad
We use the ModelNet-40 dataset \cite{wu2015modelnet} for model training, which is a widely used dataset in 3D computer vision, with a comprehensive collection of CAD models for common objects. For fair comparison, we adopt the noisy subset of the dataset provided by \cite{hermosilla2019TotalDenoising}, which consists of 15 different classes with 7 different polygonal models for each class (5 for training and 2 for testing).  Specifically, 10K to 50K points are sampled from the surface grid using Poisson disk sampling with resolutions, which are then perturbed \emph{only by Gaussian noise} with a standard deviation of 0.5\%{} to 1.5\%{} of the radius of the bounding sphere and normalized to the unit sphere before being entered into the model. Note that we did not use any clean point clouds in the datasets as in \cite{hermosilla2019TotalDenoising} to demonstrate the unsupervised nature of our method. 

For quantitative evaluation, we employ two benchmarks: the ModelNet-40 test set which includes 60 objects \cite{luo2020DMR}, and the PU-Net test set consisting of 20 objects adopted in \cite{luo_score-based_2021}. The performance of the model is evaluated using data with added Gaussian noise or a simulated LiDAR scanning noise model, specifically the Velodyne HDL-64E 3D LiDAR scanner. For the Gaussian noise case, we used Poisson disk sampling to sample point clouds from each shape at resolution levels of 10K and 50K points. The simulated LiDAR noise dataset \cite{gschwandtner2011blensor} encompasses point clouds with varying densities, ranging from 3K to 120K points, yielding a comprehensive dataset with 12 million training points and 5 million testing points.  We also evaluated the model using a real-world point cloud dataset \textit{Paris-rue-Madame}  \cite{serna2014paris}, which was obtained with a laser scanner from street scenes. 

\noindent\textbf{Model Training and Inference}\quad 
First, we perturb the noisy data to generate the training point clouds. Specifically, for each point $Y_i$ in the noisy point cloud, we generate a perturbed point $Y'_i$ by adding small Gaussian noise $\epsilon_i \sim \mathcal{N}(0, \sigma_t^2 I)$. We trained the network for 400 epochs using Adam optimizer \cite{kingma2014adam} with a learning rate of 0.0002 and a weight decay of 0.0001. During training, we anneal $\sigma_t$ linearly from time $\sigma_{\text{max}}=0.031$ to $\sigma_{\text{min}}=0.001$, which covers the range of test data. We found that the annealing procedure is important for training the model to adapt to different noise levels and to accurately estimate the score function. After training the network, we apply Tweedie's formula to estimate the clean point cloud $\mathbf{x}$. All experiments are carried out on an NVIDIA RTX 3080ti GPU with 12 GB of memory. 
%-------------------------------------------------------------------------

\noindent\textbf{Baseline}\quad We compare our method with various classical methods including bilateral filtering based (Bilateral) \cite{digne2017bilateral}, low rank based method (Lowrank) \cite{Lu2020lowrank} and graph Laplacian regularizer (GLR) \cite{zeng2019GLR}, as well as state-of-the-art unsupervised learning algorithms including Total Denoising (TotalDn) \cite{hermosilla2019TotalDenoising},  the unsupervised version of DMRDenoise (DMR-U) \cite{luo2020DMR} and the Score-Based Denoising algorithm (Score-U) \cite{luo_score-based_2021}. We do not compare with supervised learning methods as it is not the focus of the paper. \cite{noise4Wang2024} is not included as the paper does not provide a code or pre-trained model. 

\noindent\textbf{Metrics}\quad We use two commonly used metrics, the Chamfer distance (CD) \cite{cdloss} and the point-to-mesh distance (P2M) \cite{ravi2020pytorch3d}, to  evaluate model performance. Since point clouds vary in size, we normalize the denoising results to the unit sphere before computing the metric.

\subsection{Quantitative Results}
\label{sec:experiment:quantitative}
We test our trained model on three types of datasets, including the ModelNet-40 dataset/ PU-Net dataset with isotropic Gaussian noise \cite{yu2018PUNet}, and ModelNet-40 with simulated
LiDAR noise \cite{hermosilla2019TotalDenoising}. For Gaussian noises, the noise level (standard deviation of Gaussian) is set to 1\%, 2\%, and 3\% of the radius of the shape boundary sphere. Below we report the results for each case. For all results, we assume the noise parameter is unknown and is estimated following the procedure in \cref{sec:method:denosing}.

\noindent\textbf{Accuracy results}\quad \cref{table:quantitativemd} shows the comparative performance of various unsupervised denoising methods across sparse (10k) and dense (50k) point clouds with the ModelNet-40 dataset. It can be observed that Noise2Score3D consistently achieves better results than other unsupervised approaches in both Chamfer distance (CD) and point-to-mesh (P2M) metrics, and rivals GLR, the best-performing classical method. Notably, our method demonstrates consistent performance across various noise levels, particularly at high noise levels where other unsupervised learning methods may produce large errors. 

To assess the model's generalization ability, we reevaluate it on the PU-Net dataset (\cref{table:quantitativepu}). It can be seen that Noise2Score3D outperforms other unsupervised learning-based methods, highlighting its strong generalization ability across datasets. It is worth noting that our model is trained solely on the ModelNet-40 dataset, whereas the classical methods and learning-based methods are optimized/trained on the PU-Net dataset directly. We also tested the accuracy of the inference results using the true noise parameters and found that CD and P2M could be further improved (for details please refer to the Supplementary Material \cref{table:sigma}). 

We also conducted evaluations on a synthetic dataset with emulated LiDAR sensor noise \cite{hermosilla2019TotalDenoising}. The results (Table 3) show that Noise2Score3D significantly outperforms other unsupervised learning methods across various noise levels and approaches the performance of the best classical methods. Overall, our method demonstrates persistent denoising performance across various noise conditions. 

%-------------------------------------------------------------------------
% \noindent\textbf{Inference time}\quad  We compare the inference time of Noise2Score3D with other learning-based methods (TotalDn, DMR-U and Score-U). Experiments are run on the same GPU device (NVIDIA RTX 3080ti) with the exception for TotalDn, which we can only run on a NVIDIA RTX 2080Ti GPU. The classical methods are not compared here for their different implementation platforms (Matlab). We report the average inference time on 20 PU-Net point clouds with 10K or 50K sampled points.  As shown in \cref{table:time}, Noise2Score3D is, on average, 15 (50) or more times faster in inference times than other methods on 10K (50K) point clouds, confirming the efficiency of our approach.
%-------------------------------------------------------------------------

\subsection{Qualitative Results}
\label{sec:experiment:qualitative}
We visualize denoising results of the proposed method and the competing baselines under Gaussian and simulated LiDAR noise, as shown in \cref{fig:visualization}. The color of each point indicates its reconstruction error, which is measured only by the point-to-mesh distance, unlike in \cref{sec:experiment:setup}, where P2M consists of two distances: point-to-mesh and mesh-to-point. From \cref{fig:visualization}, it is evident that our results are cleaner and visually superior compared to those of other unsupervised learning methods and most classical approaches. 
Furthermore, we conduct a qualitative study on the real-world dataset \textit{Paris-rue-Madame}\cite{serna2014paris}. The results are shown in \cref{fig:rue32}, in which the left and right parts of each sub-figure are rendered from different views. Due to the varying and intricate nature of noise in point clouds, many methods struggle to maintain a balance between effective denoising and the preservation of fine details, particularly under high-conditions. Additionally, point clouds produced by these methods often suffer from nonuniform distribution, as shown in the right part of the 
subfigures in \cref{fig:rue32}. In contrast, Noise2Score3D produces cleaner and smoother results, in the meanwhile better preserving the details.

\subsection{Inference time}

We compare the inference time of Noise2Score3D with other learning-based methods (TotalDn, DMR-U and Score-U). Experiments are run on the same GPU device (NVIDIA RTX 3080ti) with the exception for TotalDn, which we can only run on a NVIDIA RTX 2080Ti GPU. The classical methods are not compared here for their different implementation platforms (Matlab). We report the average inference time on 20 PU-Net point clouds with 10K or 50K sampled points.  As shown in \cref{table:time}, Noise2Score3D is, on average, 15 (50) or more times faster in inference times than other methods on 10K (50K) point clouds, confirming the efficiency of our approach.

%-------------------------------------------------------------------------
\begin{table}[ht!]
\caption{Comparison of inference time for different algorithms on the PU-Net dataset\cite{yu2018PUNet}.}
\vspace{-0.5cm}
    \begin{center}
    \begin{tabular}{ccc}
    \toprule
                       &\multicolumn{2}{c}{Time (s)}   \\
    Dataset: \bf PU-Net & \multicolumn{1}{c}{10K, 2\%} & \multicolumn{1}{c}{50K, 2\%}  \\
    \midrule
    TotalDn\cite{hermosilla2019TotalDenoising}     & 61.83 & 6198.95   \\
    DMR-U\cite{luo2020DMR}     & 39.08 & 315.71   \\
    Score-U\cite{luo_score-based_2021}   & 33.39 & 308.42   \\
    \midrule
    \bf Ours & \bf 2.10 & \bf 5.83  \\
\bottomrule
\end{tabular}

    \end{center}
    
\label{table:time}
\end{table}

%-------------------------------------------------------------------------

\vspace{-0.5cm}
\subsection{Ablation Studies}
\label{sec:experiment:ablation}
% We performed four ablation studies:

\noindent\textbf{Choices of loss function}\quad
We compare the current results of AR-DAE loss with that of using DAE loss \cite{Alain_Bengio_dae} in \cref{table:ablation1}, showing the superiority of AR-DAE loss. Training and evaluation are both on the ModelNet-40 dataset with 50k points.

\begin{table}[ht!]
\caption{Effect of different loss function choices in training. The results are on the ModelNet-40 dataset.}
\vspace{-0.5cm}
\setlength{\tabcolsep}{2pt} 
\begin{center}
\small %
\begin{tabular}{c|cccccc}
\hline
\multicolumn{1}{c|}{Noise level}    & \multicolumn{2}{c}{1\%} & \multicolumn{2}{c}{2\%} & \multicolumn{2}{c}{3\%} \\ \cline{1-7} 

\multicolumn{1}{c|}{Loss} & CD & P2M   & CD & P2M    & CD & P2M    \\ \hline

\multicolumn{1}{c|}{DAE}  & 5.153 & 5.040      & 16.481  & 16.092      & 34.999 & 34.530      \\
\multicolumn{1}{c|}{AR-DAE(\textbf{Ours})}  & \bf1.881    & \bf1.988           & \bf4.393   & \bf4.002          & \bf8.917  & \bf8.181           \\ \hline
\end{tabular}
\end{center}
\label{table:ablation1}
\end{table}

\noindent\textbf{Annealing noise levels $\sigma_t$ in training}\quad
We compare the results of the current model trained by gradually annealing noise levels $\sigma_t$ with those of a fixed model $\sigma_t$. The results are shown in \cref{table:ablation2}, from which we can see that, while the model trained with fixed $\sigma_t$ can get better results for test data with similar noise levels, the annealing procedure enables stable denoising outcomes across different noise levels.

%-------------------------------------------------------------------------
\begin{table}[ht!]
\caption{Effect of different $\sigma_t$ choices in training.}
\vspace{-0.5cm}
\setlength{\tabcolsep}{2pt} 
\begin{center}
\small %
\begin{tabular}{cc|cccccc}
\hline
\multicolumn{1}{c|}{\multirow{2}{*}{Dataset}} & Noise level   & \multicolumn{2}{c}{1\%} & \multicolumn{2}{c}{2\%} & \multicolumn{2}{c}{3\%} \\ \cline{2-8} 

\multicolumn{1}{c|}{} &$\sigma_t$   & CD & P2M   & CD& P2M    & CD & P2M    \\ \hline

\multicolumn{1}{c|}{\multirow{4}{*}{ModelNet-40}} &0.03 & 2.880  & 2.656      & \bf4.224  & 3.966     & 8.511  & 7.804      \\

\multicolumn{1}{c|}{} &0.02 & 2.425  & 2.359      & 4.300  &\bf3.950      & \bf8.450  & \bf7.677 \\

\multicolumn{1}{c|}{} &0.01 & 2.014   & 2.078      & 4.376  & 3.997      &8.812   & 8.032      \\ 

\multicolumn{1}{c|}{} &\bf \textbf{Ours}   & \bf1.881    & \bf1.988           & 4.393   & 4.002          & 8.917  & 8.181           \\ \cline{1-8} 

\multicolumn{1}{c|}{\multirow{4}{*}{PU-Net}}& 0.03 & 1.579  & 0.762      & 2.585  & 1.576     & 2.753  & 1.706      \\
\multicolumn{1}{c|}{} & 0.02 & 1.471  & 0.675      & 2.238  &1.279      & 2.545  & 1.519    \\
\multicolumn{1}{c|}{} & 0.01 & 1.254   & 0.509      &1.711  & \bf0.847      &\bf2.395   & \bf1.378      \\ 
\multicolumn{1}{c|}{} &\bf \textbf{Ours}  &\bf0.833  & \bf0.461    & \bf1.532  & 0.970     &2.434   &1.704     \\ 
% \cline{2-8} 

% \multicolumn{1}{c|}{\bf Ours}     & \bf0.991    & \bf0.317           & \bf1.438   & \bf0.623          & 2.418  & 1.389  \\ 
\hline
\end{tabular}
\end{center}
\label{table:ablation2}
\end{table}

%-------------------------------------------------------------------------

\noindent\textbf{Result comparison using noise parameters estimated with $TV_{PC}$ vs. true values}\quad 
\cref{table:sigma} shows the results with noise parameter determined by the proposed $TV_{PC}$ versus that using the true value of $\sigma$. We observe that the metrics of denoising outcomes are similar between two cases across different noise levels, validating the proposed $TV_{PC}$ in assessing the quality of denoising result and estimating the unknown noise parameters.

%-------------------------------------------------------------------------
\begin{table}[!ht]
\caption{Denoising results using estimated noise parameters by $TV_{PC}$ ($\sigma_{tv}$)  vs. true values ($\sigma_{true}$) on the PU-Net dataset\cite{yu2018PUNet}.}
\vspace{-0.5cm}
    \begin{center}
     \resizebox{\columnwidth}{!}{%
    \begin{tabular}{ccccccc|cccccc}
    \toprule
    \bf PU-Net & \multicolumn{6}{c|}{10K} & \multicolumn{6}{c}{50K} \\
    Inference  &\multicolumn{2}{c}{1\%}   &\multicolumn{2}{c}{2\%} &\multicolumn{2}{c|}{3\%} 
              &\multicolumn{2}{c}{1\%}   &\multicolumn{2}{c}{2\%} &\multicolumn{2}{c}{3\%}   \\ \midrule
              & CD & P2M   & CD  & P2M   & CD & P2M   & CD & P2M   & CD& P2M    & CD& P2M        \\
    \midrule
    \multicolumn{1}{c|}{$\sigma_\text{true}$}
    & \bf2.806   & \bf0.661      & \bf3.868  & \bf1.173    & \bf5.192 & \bf2.228   
    & 0.991 & \bf0.317        & \bf1.438  & \bf0.623       & \bf2.418  & \bf1.389 \\
    \midrule
    
    \multicolumn{1}{c|}{$\sigma_\text{tv}$} 
    & 2.848  & 1.106           & 4.190   & 1.818         & 5.583   & 2.947           
    & \bf0.833   & 0.461          & 1.532  & 0.970          & 2.434   & 1.704  \\
\bottomrule
\end{tabular}
%-------------------------------------------------------------------------
% \begin{tabular}{cccc|ccc}
%     \toprule
%     \bf PU-Net & \multicolumn{3}{c|}{10K} & \multicolumn{3}{c}{50K}  \\
%     Inference & 1\%  & 2\%  & 3\%      & 1\%  & 2\%  & 3\%\\
%     \midrule
%     $\sigma_\text{true}$   & 0.01 & 0.02 & 0.03      & 0.01 & 0.02 & 0.03 \\
%     \midrule
%     $\sigma_\text{tv}$ &  0.0151 &  0.0215  &  0.0303     
%                 &  0.0122 &  0.0207  &  0.0285  \\
%     \midrule
%     Error(\%)  & 51\%  & 7.5\% & 1\%   & 22\%  & 3.5\% & 5\%      \\ 
% \bottomrule
% \end{tabular}
%-------------------------------------------------------------------------
 }%
    \end{center}
    \label{table:sigma}
\end{table}

%-------------------------------------------------------------------------

\section{Conclusions}
Despite significant research efforts, the denoising of 3D point clouds remains a challenging problem. The proposed Noise2Score3D framework introduces a novel and effective Bayesian approach for unsupervised point cloud denoising. Noise2Score3D achieves SOTA denoising performance among unsupervised learning methods, while being highly efficient. It also shows strong generalization ability across varying datasets and noise conditions. Moreover, our proposed $TV_{PC}$ metric enables estimation of unknown noise parameters, greatly enhancing the method's applicability in real-world scenarios.

The current method is restricted to exponential family noises, and assumes an uniform noise level across all points in the cloud. But it is possible to be extended to more complex noise models, such as multiplicative and structurally correlated noise\cite{scoreXie2024}. 
% Also the current inference process assumes an uniform noise level, which could be
Future work includes extending the method to other input types, e.g. color point clouds and more realistic noise setup. Another direction is to apply the score estimation framework to other tasks, e.g. point cloud upsampling and completion. We believe that our approach can inspire future research in the application of Bayesian methods in point cloud denoising and, more generally, 3D data processing.

\section*{Acknowledgement}
The work was supported by Guangdong Provincial Quantum Science Strategic Initiative (GDZX2306001, GDZX2303001) the startup fund of Shenzhen City, Shenzhen Fundamental research project (Grant No. JCYJ20241202123903005).

% \clearpage
{
    \small
    \bibliographystyle{ieeenat_fullname}
    \bibliography{main}

\begin{thebibliography}{57}
\providecommand{\natexlab}[1]{#1}
\providecommand{\url}[1]{\texttt{#1}}
\expandafter\ifx\csname urlstyle\endcsname\relax
  \providecommand{\doi}[1]{doi: #1}\else
  \providecommand{\doi}{doi: \begingroup \urlstyle{rm}\Url}\fi

\bibitem[Alain and Bengio(2014{\natexlab{a}})]{Alain_Bengio_dae}
Guillaume Alain and Yoshua Bengio.
\newblock What regularized auto-encoders learn from the data-generating
  distribution.
\newblock \emph{Journal of Machine Learning Research}, 15\penalty0
  (110):\penalty0 3743--3773, 2014{\natexlab{a}}.

\bibitem[Alain and Bengio(2014{\natexlab{b}})]{alain2014regularized}
Guillaume Alain and Yoshua Bengio.
\newblock What regularized auto-encoders learn from the data-generating
  distribution.
\newblock \emph{J. Mach. Learn. Res.}, 15\penalty0 (1):\penalty0 3563–3593,
  2014{\natexlab{b}}.

\bibitem[Alexa et~al.(2001)Alexa, Behr, Cohen-Or, Fleishman, Levin, and
  Silva]{pointsetsurfaces2001}
M. Alexa, J. Behr, D. Cohen-Or, S. Fleishman, D. Levin, and C.T. Silva.
\newblock Point set surfaces.
\newblock In \emph{Proceedings Visualization, 2001. VIS '01.}, pages 21--29,
  537, 2001.

\bibitem[Alexa et~al.(2003)Alexa, Behr, Cohen-Or, Fleishman, Levin, and
  Silva]{computingpointset2003}
M. Alexa, J. Behr, D. Cohen-Or, S. Fleishman, D. Levin, and C.T. Silva.
\newblock Computing and rendering point set surfaces.
\newblock \emph{IEEE Transactions on Visualization and Computer Graphics},
  9\penalty0 (1):\penalty0 3--15, 2003.

\bibitem[Amenta and Kil(2004)]{definingpointset2004}
Nina Amenta and Yong~Joo Kil.
\newblock Defining point-set surfaces.
\newblock \emph{ACM Trans. Graph.}, 23\penalty0 (3):\penalty0 264–270, 2004.

\bibitem[Avron et~al.(2010)Avron, Sharf, Greif, and Cohen-Or]{l1sparse2010}
Haim Avron, Andrei Sharf, Chen Greif, and Daniel Cohen-Or.
\newblock $\ell$1-sparse reconstruction of sharp point set surfaces.
\newblock \emph{ACM Trans. Graph.}, 29\penalty0 (5), 2010.

\bibitem[Batson and Royer(2019)]{2019Noise2SelfBD}
Joshua Batson and Loic Royer.
\newblock {N}oise2{S}elf: Blind denoising by self-supervision.
\newblock In \emph{Proceedings of the 36th International Conference on Machine
  Learning}, pages 524--533. PMLR, 2019.

\bibitem[Cai et~al.(2024)Cai, Choi, Li, and Yin]{doi:10.1137/23M1614456}
Jian-Feng Cai, Jae~Kyu Choi, Jingyang Li, and Guojian Yin.
\newblock Restoration guarantee of image inpainting via low rank patch matrix
  completion.
\newblock \emph{SIAM Journal on Imaging Sciences}, 17\penalty0 (3):\penalty0
  1879--1908, 2024.

\bibitem[Cai et~al.(2020)Cai, Yang, Averbuch-Elor, Hao, Belongie, Snavely, and
  Hariharan]{cai2020learning}
Ruojin Cai, Guandao Yang, Hadar Averbuch-Elor, Zekun Hao, Serge Belongie, Noah
  Snavely, and Bharath Hariharan.
\newblock Learning gradient fields for shape generation.
\newblock In \emph{Computer Vision--ECCV 2020: 16th European Conference,
  Glasgow, UK, August 23--28, 2020, Proceedings, Part III 16}, pages 364--381.
  Springer, 2020.

\bibitem[Chen et~al.(2022)Chen, Luo, Hu, et~al.]{chen2022deep}
Haolan Chen, Shitong Luo, Wei Hu, et~al.
\newblock Deep point set resampling via gradient fields.
\newblock \emph{IEEE Transactions on Pattern Analysis and Machine
  Intelligence}, 45\penalty0 (3):\penalty0 2913--2930, 2022.

\bibitem[de~Silva~Edirimuni et~al.(2023)de~Silva~Edirimuni, Lu, Shao, Li,
  Robles-Kelly, and He]{de_Silva_Edirimuni_2023_CVPR}
Dasith de Silva~Edirimuni, Xuequan Lu, Zhiwen Shao, Gang Li, Antonio
  Robles-Kelly, and Ying He.
\newblock Iterativepfn: True iterative point cloud filtering.
\newblock In \emph{Proceedings of the IEEE/CVF Conference on Computer Vision
  and Pattern Recognition (CVPR)}, pages 13530--13539, 2023.

\bibitem[Digne and De~Franchis(2017)]{digne2017bilateral}
Julie Digne and Carlo De~Franchis.
\newblock The bilateral filter for point clouds.
\newblock \emph{Image Processing On Line}, 7:\penalty0 278--287, 2017.

\bibitem[Duan et~al.(2018)Duan, Chen, and Kovacevic]{multi-projection2018duan}
Chaojing Duan, Siheng Chen, and Jelena Kovacevic.
\newblock Weighted multi-projection: 3d point cloud denoising with tangent
  planes.
\newblock In \emph{2018 IEEE Global Conference on Signal and Information
  Processing (GlobalSIP)}, pages 725--729, 2018.

\bibitem[Duan et~al.(2019)Duan, Chen, and Kovacevic]{NPD2019}
Chaojing Duan, Siheng Chen, and Jelena Kovacevic.
\newblock 3d point cloud denoising via deep neural network based local surface
  estimation.
\newblock In \emph{ICASSP 2019 - 2019 IEEE International Conference on
  Acoustics, Speech and Signal Processing (ICASSP)}, pages 8553--8557, 2019.

\bibitem[Efron(2011)]{efron2011tweedie}
Bradley Efron.
\newblock Tweedie’s formula and selection bias.
\newblock \emph{Journal of the American Statistical Association}, 106:\penalty0
  1602--1614, 2011.

\bibitem[Fan et~al.(2017)Fan, Su, and Guibas]{cdloss}
Haoqiang Fan, Hao Su, and Leonidas~J Guibas.
\newblock A point set generation network for 3d object reconstruction from a
  single image.
\newblock In \emph{Proceedings of the IEEE conference on computer vision and
  pattern recognition}, pages 605--613, 2017.

\bibitem[Fleishman et~al.(2005)Fleishman, Cohen-Or, and
  Silva]{Robustmoving2005}
Shachar Fleishman, Daniel Cohen-Or, and Cl\'{a}udio~T. Silva.
\newblock Robust moving least-squares fitting with sharp features.
\newblock \emph{ACM Trans. Graph.}, 24\penalty0 (3):\penalty0 544–552, 2005.

\bibitem[Gravel et~al.(2004)Gravel, Beaudoin, and De~Guise]{gravel2004method}
Pierre Gravel, Gilles Beaudoin, and Jacques~A De~Guise.
\newblock A method for modeling noise in medical images.
\newblock \emph{IEEE Transactions on medical imaging}, 23\penalty0
  (10):\penalty0 1221--1232, 2004.

\bibitem[Gschwandtner et~al.(2011)Gschwandtner, Kwitt, Uhl, and
  Pree]{gschwandtner2011blensor}
Michael Gschwandtner, Roland Kwitt, Andreas Uhl, and Wolfgang Pree.
\newblock Blensor: Blender sensor simulation toolbox.
\newblock In \emph{International Symposium on Visual Computing}, pages
  199--208. Springer, 2011.

\bibitem[Guerrero et~al.(2017)Guerrero, Kleiman, Ovsjanikov, and
  Mitra]{Guerrero2017PCPNetLL}
Paul Guerrero, Yanir Kleiman, Maks Ovsjanikov, and Niloy~Jyoti Mitra.
\newblock Pcpnet learning local shape properties from raw point clouds.
\newblock \emph{Computer Graphics Forum}, 37, 2017.

\bibitem[Hermosilla et~al.(2019)Hermosilla, Ritschel, and
  Ropinski]{hermosilla2019TotalDenoising}
Pedro Hermosilla, Tobias Ritschel, and Timo Ropinski.
\newblock Total denoising: Unsupervised learning of 3d point cloud cleaning.
\newblock In \emph{Proceedings of the IEEE International Conference on Computer
  Vision}, pages 52--60, 2019.

\bibitem[Huang et~al.(2009)Huang, Li, Zhang, Ascher, and Cohen-Or]{wlop2009HH}
Hui Huang, Dan Li, Hao Zhang, Uri Ascher, and Daniel Cohen-Or.
\newblock Consolidation of unorganized point clouds for surface reconstruction.
\newblock \emph{ACM Trans. Graph.}, 28\penalty0 (5):\penalty0 1–7, 2009.

\bibitem[Huang et~al.(2021)Huang, Li, Jia, Lu, and
  Liu]{huang2021neighbor2neighbor}
Tao Huang, Songjiang Li, Xu Jia, Huchuan Lu, and Jianzhuang Liu.
\newblock Neighbor2neighbor: Self-supervised denoising from single noisy
  images.
\newblock In \emph{2021 IEEE/CVF Conference on Computer Vision and Pattern
  Recognition (CVPR)}, pages 14776--14785, 2021.

\bibitem[Hyvärinen(2005)]{hyvarinen2005estimation}
Aapo Hyvärinen.
\newblock Estimation of non-normalized statistical models by score matching.
\newblock \emph{Journal of Machine Learning Research}, 6:\penalty0 695--709,
  2005.

\bibitem[Kim and Ye(2021{\natexlab{a}})]{kim_noise2score_2021}
Kwanyoung Kim and Jong-Chul Ye.
\newblock Noise2score: Tweedie's approach to self-supervised image denoising
  without clean images.
\newblock In \emph{Neural Information Processing Systems}, 2021{\natexlab{a}}.

\bibitem[Kim and Ye(2021{\natexlab{b}})]{noise2score2021}
Kwanyoung Kim and Jong~Chul Ye.
\newblock Noise2score: Tweedie’s approach to self-supervised image denoising
  without clean images.
\newblock In \emph{Advances in Neural Information Processing Systems}, pages
  864--874. Curran Associates, Inc., 2021{\natexlab{b}}.

\bibitem[Kingma(2014)]{kingma2014adam}
Diederik~P Kingma.
\newblock Adam: A method for stochastic optimization.
\newblock \emph{arXiv preprint arXiv:1412.6980}, 2014.

\bibitem[Krull et~al.(2018)Krull, Buchholz, and Jug]{2018Noise2VoidL}
Alexander Krull, Tim-Oliver Buchholz, and Florian Jug.
\newblock Noise2void - learning denoising from single noisy images.
\newblock \emph{2019 IEEE/CVF Conference on Computer Vision and Pattern
  Recognition (CVPR)}, pages 2124--2132, 2018.

\bibitem[Lai et~al.(2019)Lai, Huang, Ahuja, and Yang]{Charbonnier}
Wei-Sheng Lai, Jia-Bin Huang, Narendra Ahuja, and Ming-Hsuan Yang.
\newblock Fast and accurate image super-resolution with deep laplacian pyramid
  networks.
\newblock \emph{IEEE Transactions on Pattern Analysis and Machine
  Intelligence}, 41\penalty0 (11):\penalty0 2599--2613, 2019.

\bibitem[Lehtinen et~al.(2018)Lehtinen, Munkberg, Hasselgren, Laine, Karras,
  Aittala, and Aila]{2018Noise2NoiseLI}
Jaakko Lehtinen, Jacob Munkberg, Jon Hasselgren, Samuli Laine, Tero Karras,
  Miika Aittala, and Timo Aila.
\newblock Noise2noise: Learning image restoration without clean data.
\newblock \emph{ArXiv}, abs/1803.04189, 2018.

\bibitem[Lim et~al.(2020)Lim, Courville, Pal, and Huang]{lim2020ar-dae}
Jae~Hyun Lim, Aaron Courville, Christopher Pal, and Chin-Wei Huang.
\newblock {AR}-{DAE}: Towards unbiased neural entropy gradient estimation.
\newblock In \emph{Proceedings of the 37th International Conference on Machine
  Learning}, pages 6061--6071. PMLR, 2020.

\bibitem[Liu et~al.(2024)Liu, Zhao, Zhan, Liu, Chen, and He]{10173632}
Zheng Liu, Yaowu Zhao, Sijing Zhan, Yuanyuan Liu, Renjie Chen, and Ying He.
\newblock Pcdnf: Revisiting learning-based point cloud denoising via joint
  normal filtering.
\newblock \emph{IEEE Transactions on Visualization and Computer Graphics},
  30\penalty0 (8):\penalty0 5419--5436, 2024.

\bibitem[Lu et~al.(2020)Lu, Schaefer, Luo, Ma, and He]{Lu2020lowrank}
Xuequan Lu, Scott Schaefer, Jun Luo, Lizhuang Ma, and Ying He.
\newblock Low rank matrix approximation for 3d geometry filtering.
\newblock \emph{IEEE Transactions on Visualization and Computer Graphics},
  pages 1--1, 2020.

\bibitem[Luisier et~al.(2010)Luisier, Blu, and Unser]{luisier2010image}
Florian Luisier, Thierry Blu, and Michael Unser.
\newblock Image denoising in mixed poisson--gaussian noise.
\newblock \emph{IEEE Transactions on image processing}, 20\penalty0
  (3):\penalty0 696--708, 2010.

\bibitem[Luo and Hu(2020)]{luo2020DMR}
Shitong Luo and Wei Hu.
\newblock Differentiable manifold reconstruction for point cloud denoising.
\newblock In \emph{Proceedings of the 28th ACM International Conference on
  Multimedia}, pages 1330--1338, 2020.

\bibitem[Luo and Hu(2021)]{luo_score-based_2021}
Shitong Luo and Wei Hu.
\newblock Score-based point cloud denoising.
\newblock In \emph{2021 IEEE/CVF International Conference on Computer Vision
  (ICCV)}, pages 4563--4572, 2021.

\bibitem[Preiner et~al.(2014)Preiner, Mattausch, Arikan, Pajarola, and
  Wimmer]{clop2014PR}
Reinhold Preiner, Oliver Mattausch, Murat Arikan, Renato Pajarola, and Michael
  Wimmer.
\newblock Continuous projection for fast l1 reconstruction.
\newblock \emph{ACM Trans. Graph.}, 33\penalty0 (4), 2014.

\bibitem[Qi et~al.(2017{\natexlab{a}})Qi, Su, Mo, and Guibas]{qi2017pointnet}
Charles~R Qi, Hao Su, Kaichun Mo, and Leonidas~J Guibas.
\newblock Pointnet: Deep learning on point sets for 3d classification and
  segmentation.
\newblock In \emph{Proceedings of the IEEE conference on computer vision and
  pattern recognition}, pages 652--660, 2017{\natexlab{a}}.

\bibitem[Qi et~al.(2017{\natexlab{b}})Qi, Yi, Su, and Guibas]{qi2017pointnet2}
Charles~Ruizhongtai Qi, Li Yi, Hao Su, and Leonidas~J Guibas.
\newblock Pointnet++: Deep hierarchical feature learning on point sets in a
  metric space.
\newblock In \emph{Advances in neural information processing systems}, pages
  5099--5108, 2017{\natexlab{b}}.

\bibitem[Quan et~al.(2024)Quan, Yu, Nie, Wang, Feng, An, and
  Yang]{quan2024deep}
Siwen Quan, Junhao Yu, Ziming Nie, Muze Wang, Sijia Feng, Pei An, and Jiaqi
  Yang.
\newblock Deep learning for 3d point cloud enhancement: A survey.
\newblock \emph{arXiv preprint arXiv:2411.00857}, 2024.

\bibitem[Rakotosaona et~al.(2020)Rakotosaona, La~Barbera, Guerrero, Mitra, and
  Ovsjanikov]{rakotosaona2020PCN}
Marie-Julie Rakotosaona, Vittorio La~Barbera, Paul Guerrero, Niloy~J Mitra, and
  Maks Ovsjanikov.
\newblock Pointcleannet: Learning to denoise and remove outliers from dense
  point clouds.
\newblock In \emph{Computer Graphics Forum}, pages 185--203. Wiley Online
  Library, 2020.

\bibitem[Ravi et~al.(2020)Ravi, Reizenstein, Novotny, Gordon, Lo, Johnson, and
  Gkioxari]{ravi2020pytorch3d}
Nikhila Ravi, Jeremy Reizenstein, David Novotny, Taylor Gordon, Wan-Yen Lo,
  Justin Johnson, and Georgia Gkioxari.
\newblock Accelerating 3d deep learning with pytorch3d.
\newblock \emph{arXiv:2007.08501}, 2020.

\bibitem[Rudin et~al.(1992)Rudin, Osher, and Fatemi]{RUDIN1992}
Leonid~I. Rudin, Stanley Osher, and Emad Fatemi.
\newblock Nonlinear total variation based noise removal algorithms.
\newblock \emph{Physica D: Nonlinear Phenomena}, 60:\penalty0 259--268, 1992.

\bibitem[Serna et~al.(2014)Serna, Marcotegui, Goulette, and
  Deschaud]{serna2014paris}
Andr\'{e}s Serna, Beatriz Marcotegui, Fran\c{c}ois Goulette, and Jean-Emmanuel
  Deschaud.
\newblock Paris-rue-madame database.
\newblock In \emph{Proceedings of the 3rd International Conference on Pattern
  Recognition Applications and Methods}, page 819–824, Setubal, PRT, 2014.
  SCITEPRESS - Science and Technology Publications, Lda.

\bibitem[Soltanayev and Chun(2018)]{SURE2018}
Shakarim Soltanayev and Se~Young Chun.
\newblock Sure.
\newblock In \emph{Advances in Neural Information Processing Systems}, page
  3261–3271, Red Hook, NY, USA, 2018. Curran Associates Inc.

\bibitem[Song and Ermon(2019)]{song2019generative}
Yang Song and Stefano Ermon.
\newblock Generative modeling by estimating gradients of the data distribution.
\newblock \emph{Advances in neural information processing systems}, 32, 2019.

\bibitem[Thomas et~al.(2019)Thomas, Qi, Deschaud, Marcotegui, Goulette, and
  Guibas]{KPconv}
H. Thomas, C.~R. Qi, J. Deschaud, B. Marcotegui, F. Goulette, and L. Guibas.
\newblock Kpconv: Flexible and deformable convolution for point clouds.
\newblock In \emph{2019 IEEE/CVF International Conference on Computer Vision
  (ICCV)}, pages 6410--6419. IEEE Computer Society, 2019.

\bibitem[Vincent(2011)]{vincent2011connection}
Pascal Vincent.
\newblock A connection between score matching and denoising autoencoders.
\newblock \emph{Neural Computation}, 23\penalty0 (7):\penalty0 1661--1674,
  2011.

\bibitem[Wang et~al.(2024)Wang, Liu, Zhou, Wei, Deng, Murshed, and
  Lu]{noise4Wang2024}
Weijia Wang, Xiao Liu, Hailing Zhou, Lei Wei, Zhigang Deng, Manzur Murshed, and
  Xuequan Lu.
\newblock Noise4denoise: Leveraging noise for unsupervised point cloud
  denoising.
\newblock \emph{Computational Visual Media}, 10\penalty0 (4):\penalty0
  659--669, 2024.

\bibitem[Wang et~al.(2019)Wang, Sun, Liu, Sarma, Bronstein, and
  Solomon]{wang2019dynamic}
Yue Wang, Yongbin Sun, Ziwei Liu, Sanjay~E Sarma, Michael~M Bronstein, and
  Justin~M Solomon.
\newblock Dynamic graph cnn for learning on point clouds.
\newblock \emph{ACM Transactions on Graphics (TOG)}, 38\penalty0 (5):\penalty0
  1--12, 2019.

\bibitem[Wu et~al.(2015)Wu, Song, Khosla, Yu, Zhang, Tang, and
  Xiao]{wu2015modelnet}
Zhirong Wu, Shuran Song, Aditya Khosla, Fisher Yu, Linguang Zhang, Xiaoou Tang,
  and Jianxiong Xiao.
\newblock 3d shapenets: A deep representation for volumetric shapes.
\newblock In \emph{Proceedings of the IEEE conference on computer vision and
  pattern recognition}, pages 1912--1920, 2015.

\bibitem[Xie et~al.(2024)Xie, Yuan, Dong, and Li]{scoreXie2024}
Yutong Xie, Mingze Yuan, Bin Dong, and Quanzheng Li.
\newblock Unsupervised image denoising with score function.
\newblock In \emph{Proceedings of the 37th International Conference on Neural
  Information Processing Systems}, Red Hook, NY, USA, 2024. Curran Associates
  Inc.

\bibitem[Xu et~al.(2015)Xu, Zhang, Zuo, Zhang, and Feng]{xu2015patch}
Jun Xu, Lei Zhang, Wangmeng Zuo, David Zhang, and Xiangchu Feng.
\newblock Patch group based nonlocal self-similarity prior learning for image
  denoising.
\newblock In \emph{Proceedings of the IEEE international conference on computer
  vision}, pages 244--252, 2015.

\bibitem[Yang et~al.(2023)Yang, Zhang, Song, Hong, Xu, Zhao, Zhang, Cui, and
  Yang]{yang2023diffusion}
Ling Yang, Zhilong Zhang, Yang Song, Shenda Hong, Runsheng Xu, Yue Zhao, Wentao
  Zhang, Bin Cui, and Ming-Hsuan Yang.
\newblock Diffusion models: A comprehensive survey of methods and applications.
\newblock \emph{ACM Computing Surveys}, 56\penalty0 (4):\penalty0 1--39, 2023.

\bibitem[Yu et~al.(2018)Yu, Li, Fu, Cohen-Or, and Heng]{yu2018PUNet}
Lequan Yu, Xianzhi Li, Chi-Wing Fu, Daniel Cohen-Or, and Pheng-Ann Heng.
\newblock Pu-net: Point cloud upsampling network.
\newblock In \emph{Proceedings of the IEEE Conference on Computer Vision and
  Pattern Recognition}, pages 2790--2799, 2018.

\bibitem[Zeng et~al.(2019)Zeng, Cheung, Ng, Pang, and Cheng]{zeng2019GLR}
Jin Zeng, Gene Cheung, Michael Ng, Jiahao Pang, and Yang Cheng.
\newblock 3{D} point cloud denoising using graph {Laplacian} regularization of
  a low dimensional manifold model.
\newblock \emph{IEEE Transactions on Image Processing}, 29:\penalty0
  3474--3489, 2019.

\bibitem[Zhou et~al.(2022)Zhou, Sun, Li, Li, and Su]{zhou2022point}
Lang Zhou, Guoxing Sun, Yong Li, Weiqing Li, and Zhiyong Su.
\newblock Point cloud denoising review: from classical to deep learning-based
  approaches.
\newblock \emph{Graphical Models}, 121:\penalty0 101140, 2022.

\end{thebibliography}
}
\clearpage
\renewcommand\thesection{\Alph{section}}
\renewcommand\thesubsection{\thesection.\arabic{subsection}}
\setcounter{page}{1}
\setcounter{section}{0}
\maketitlesupplementary

\section{Denoising and Estimating unknown noise parameters with Total Variation for Point cloud}
\label{sec:TV}

In image analysis, Total Variation (TV) is primarily used as a quality metric to evaluate the denoising result by calculating the remaining high-frequency content \cite{RUDIN1992}. A lower TV value indicates a smoother output with fewer irregularities, whereas a higher TV value suggests the presence of residual noise or structural inconsistencies. In TV regularization, the optimization process seeks to minimize the total variation of an image, reducing sharp pixel intensity changes that correspond to noise, while preserving significant transitions corresponding to image edges. This has inspired us to propose similar metrics for point clouds.

\subsection{Definition of Total Variation for Point Cloud}

In this section, we extend the idea of minimizing gradient magnitudes to point cloud data, where the goal is to reduce noise by minimizing the geometric differences between points and their neighbors. In particular,  Total Variation for Point Cloud  ($TV_{PC}$) is defined as: 
\begin{equation} 
\label{eq:TVPC}
TV_{PC} = \sum_{i=1}^N \sum_{j \in \text{neighbors}(i)} w_{i,j} \cdot \sqrt{\| \mathbf{p}_i - \mathbf{p}_j \|^2 + \epsilon^2}
\end{equation}
where $N$ is the total number of points, \( w_{i,j} \) represents the weight between point \( \mathbf{p}_i \) and its neighbor \( \mathbf{p}_j \), $\epsilon$ is a small positive constant.
$\sqrt{\| \mathbf{p} \|^2 + \epsilon^2}$ in Eq. (\ref{eq:TVPC}) is a smooth approximation of the \(L_1\) norm, which is introduced in \cite{Charbonnier} to handle outliers and maintain robustness. Here, $\epsilon$  helps to smooth the variations and prevent over-penalization of boundary outliers, ensuring numerical stability.
For simplicity we set \( w_{i,j} \) to a constant in our method. Another choice for \( w_{i,j} \) is 
\begin{equation} 
w_{i,j} = \exp\left(-\frac{\| \mathbf{p}_i - \mathbf{p}_j \|^2}{2\sigma^2}\right),
\end{equation}
with \( \sigma \) controlling the scale of the Gaussian kernel \cite{chen2022deep} .

This is an analogy to the difference of pixel intensity in TV definition for images. Also we note that $TV_{PC}$ is similar to the Graph Laplacian Regularizer \cite{zeng2019GLR} but with smoothed \(L_1\) loss. 

\subsection{Validation of $TV_{PC}$ in estimating the noise level}
\begin{figure}
\begin{center}
\includegraphics[width=0.5\textwidth]{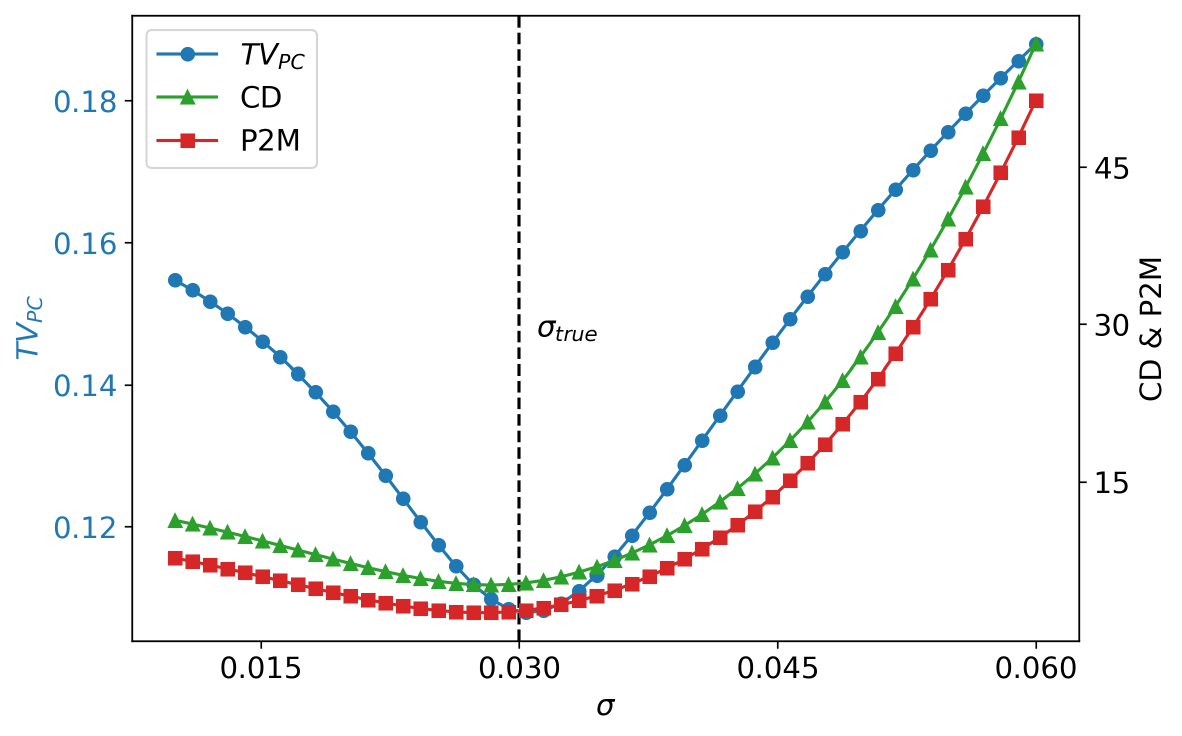}
\end{center}
\vspace{-0.6cm}
\caption{Change of average $TV_{PC}$, $CD$ and $P2M$ values with noise parameter $\sigma$ on the PU-Net dataset\cite{yu2018PUNet} .}
\label{fig:tvpc-cd-p2m}
\end{figure}

\cref{fig:tvpc-cd-p2m} shows the $TV_{PC}$, $CD$ and $P2M$ values with different noise parameter $\sigma$ for the inference result on PU-Net dataset with $\sigma$=0.03. We can observe that $TV_{PC}$ follows a similar trend 
as $CD$ and $P2M$ and reaches a minimal around the true $\sigma$ value. Note that the clean point clouds are not needed for calculation of $TV_{PC}$. This validates the use of $TV_{PC}$ in estimating unknown noise parameters for real world datasets.

%-------------------------------------------------------------------------

% \subsection{Results of Gamma noise model}
% While our current results focus on Gaussian and simulated LiDAR noise, we acknowledge the need for broader validation. To address this request, we conducted a preliminary experiment on a synthetic dataset with multiplicative Gamma noise using PU-Net dataset. The Gamma noise model is defined as:
% \vspace{-0.5em}
% \begin{equation}
% \label{eq:gamma}
% y=xn,   p(n; \alpha, \beta) = \frac{\beta^\alpha}{\Gamma(\alpha)} n^{\alpha-1} \exp(-\beta n).
% \end{equation}
% Applying Tweedie’s formula leads to a revised posterior mean \( E_{p(\mathbf{x}\mid\mathbf{y})}(x) = \beta y /(\alpha-1)-y l'(y) \). The results are summarized in R\cref{tab:gamma_noise}, where the noise parameters are chosen as \(\alpha =\beta = k\) to satisfy \(mean(n)=1\). This suggests that our method adapts to non-Gaussian noise, aligning with the theoretical extensibility of Tweedie’s formula. 

% \vspace{-1.0em}
% \begin{table}[ht!]
%     \setlength{\tabcolsep}{2pt} % 设置列间距
%     \begin{center}
%     \small %
%     \input{tables/gamma_denoise}
%     \end{center}
%     \vspace{-0.6cm}
%     \caption{Denoising results on the PU-Net 10k dataset with multiplicative Gamma noise.}
% \label{tab:gamma_noise}
% \end{table}
% \vspace{-1.0em}

\section{Performance Comparison and Ablation Study of Point Cloud Backbones}
\cref{tab:backbone} evaluates the effectiveness of different point cloud backbones,the analysis reveals that the KPConv backbone does not inherently improve the performance of DMR or Score-U, suggesting that the superior results of our method are not solely attributable to the KPConv network architecture but rather to our novel Bayesian approach. Additionally, we show the result using PointNet++ backbone, which underperforms in sparse point cloud regions and under higher noise conditions compared to KPConv, reinforcing the appropriateness of our chosen backbone for the proposed framework.

\begin{table}[ht!]
\caption{Results on the PU-Net 10k dataset (results of 50k dataset are not included due to space limitation). Results with * are taken from \cref{table:quantitativepu}. in the main text for comparison purpose.}
\vspace{-0.5cm}
    \setlength{\tabcolsep}{2pt} % 设置列间距
    \begin{center}
    \footnotesize %
    \begin{tabular}{cc|cccccc}
\hline
\multicolumn{2}{c|}{Noise level}    & \multicolumn{2}{c}{1\%} & \multicolumn{2}{c}{2\%} & \multicolumn{2}{c}{3\%} \\ \cline{1-8} 

\multicolumn{1}{c|}{Dataset} & Model   & CD & P2M   & CD& P2M    & CD & P2M    \\ \hline

\multicolumn{1}{c|}{\multirow{6}{*}{PU-Net}}  &\textbf{DMR-U$_{kpconv}$}   & 33.89      & 28.16      & 25.53      & 20.68      & 20.69       & 15.99  \\

\multicolumn{1}{c|}{}   & *DMR-U   & 5.313      & 2.522      & 6.455      & 3.317      & 8.134       & 4.647      \\\cline{2-8}

\multicolumn{1}{c|}{}     & \textbf{Score-u$_{kpconv}$}     & 3.494  & \underline{1.077}      & 4.928  &1.909      & 7.250  & 3.596    \\

\multicolumn{1}{c|}{}    & *Score-U & \underline{3.107}      & \bf0.888      & 4.675      & \underline{1.829}      & 7.225       & 3.762  \\\cline{2-8}

\multicolumn{1}{c|}{}    & \textbf{Ours$_{PointNet++}$} & 3.689     & 1.287      & 7.938      & 4.370      & 13.382       & 9.028  \\ 

\multicolumn{1}{c|}{}    & \bf *Ours       &\bf2.848  & 1.106      & \bf4.190  & \bf1.818    &\bf5.583   &\bf2.947 
           \\ \hline
\end{tabular}
    \end{center}
    
\label{tab:backbone}
\end{table}

\clearpage
\onecolumn
\section{Additional visualization results}
\label{sec:add_results}
\begin{figure*}[!ht]
\begin{center}
    \includegraphics[width=0.8\textwidth]{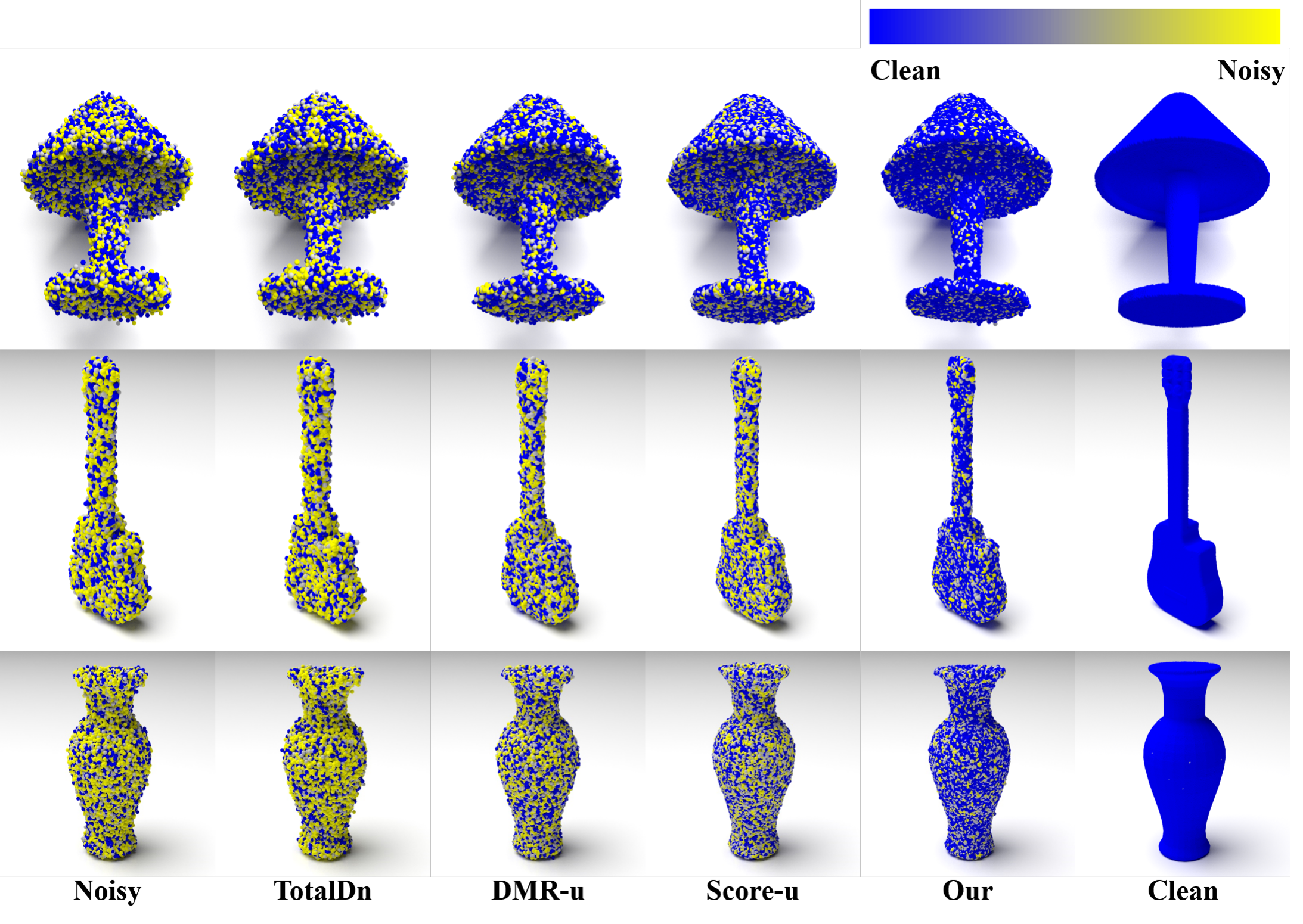}
\end{center}
\vspace{-1.0em}
\caption{Additional visualization results of different algorithms on ModelNet-40 dataset with Gaussian noise. The noise level is set to 1\%.}
\label{fig:suppl_modelnet1}
\end{figure*}

\begin{figure*}[!ht]
\begin{center}
    \includegraphics[width=0.8\textwidth]{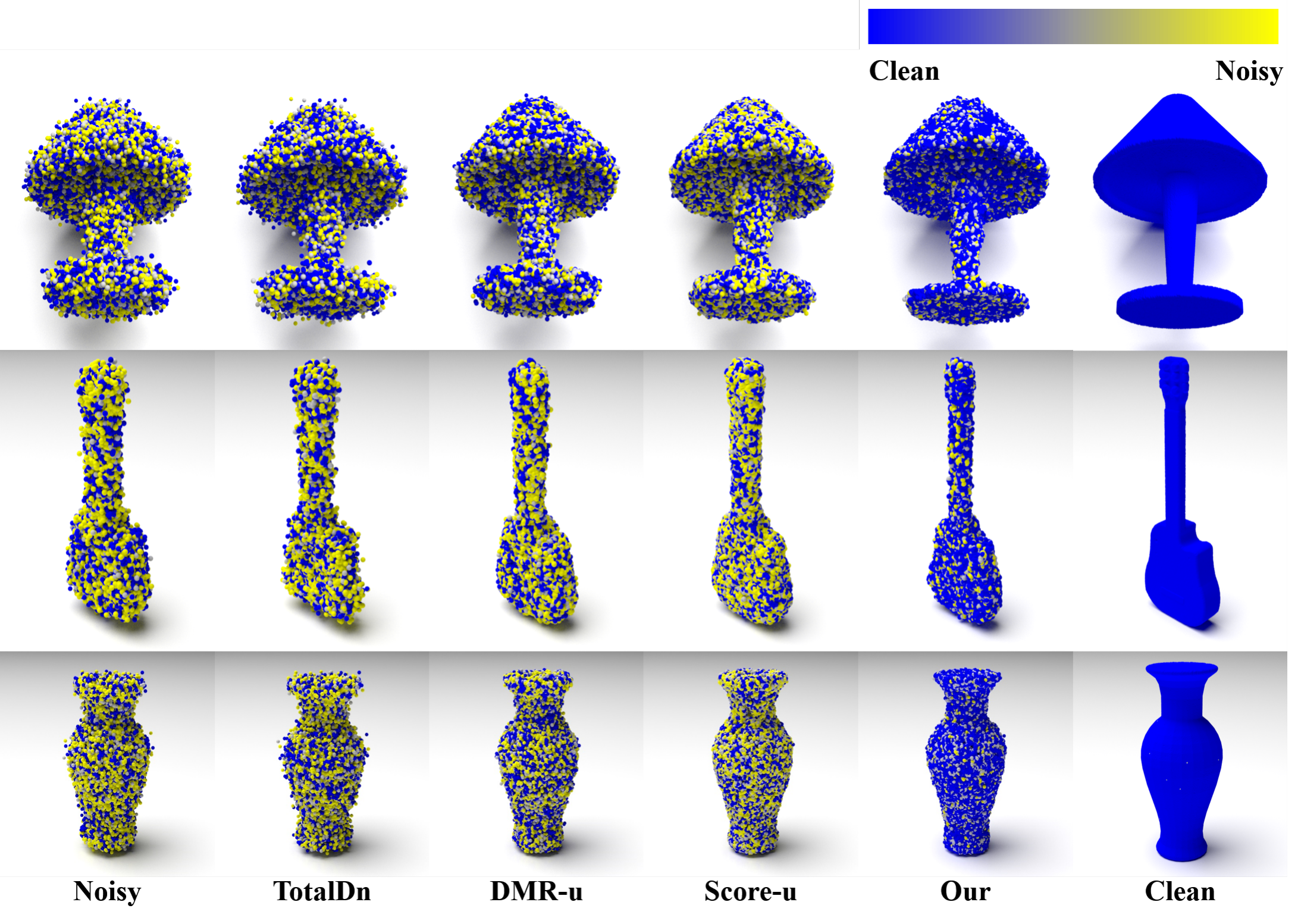}
\end{center}
\vspace{-1.0em}
\caption{Additional visualization results of different algorithms on ModelNet-40 with Gaussian noise. The noise level is set to 2\%.}
\label{fig:suppl_modelnet2}
\end{figure*}

\begin{figure*}[!ht]
\begin{center}
    \includegraphics[width=0.8\textwidth]{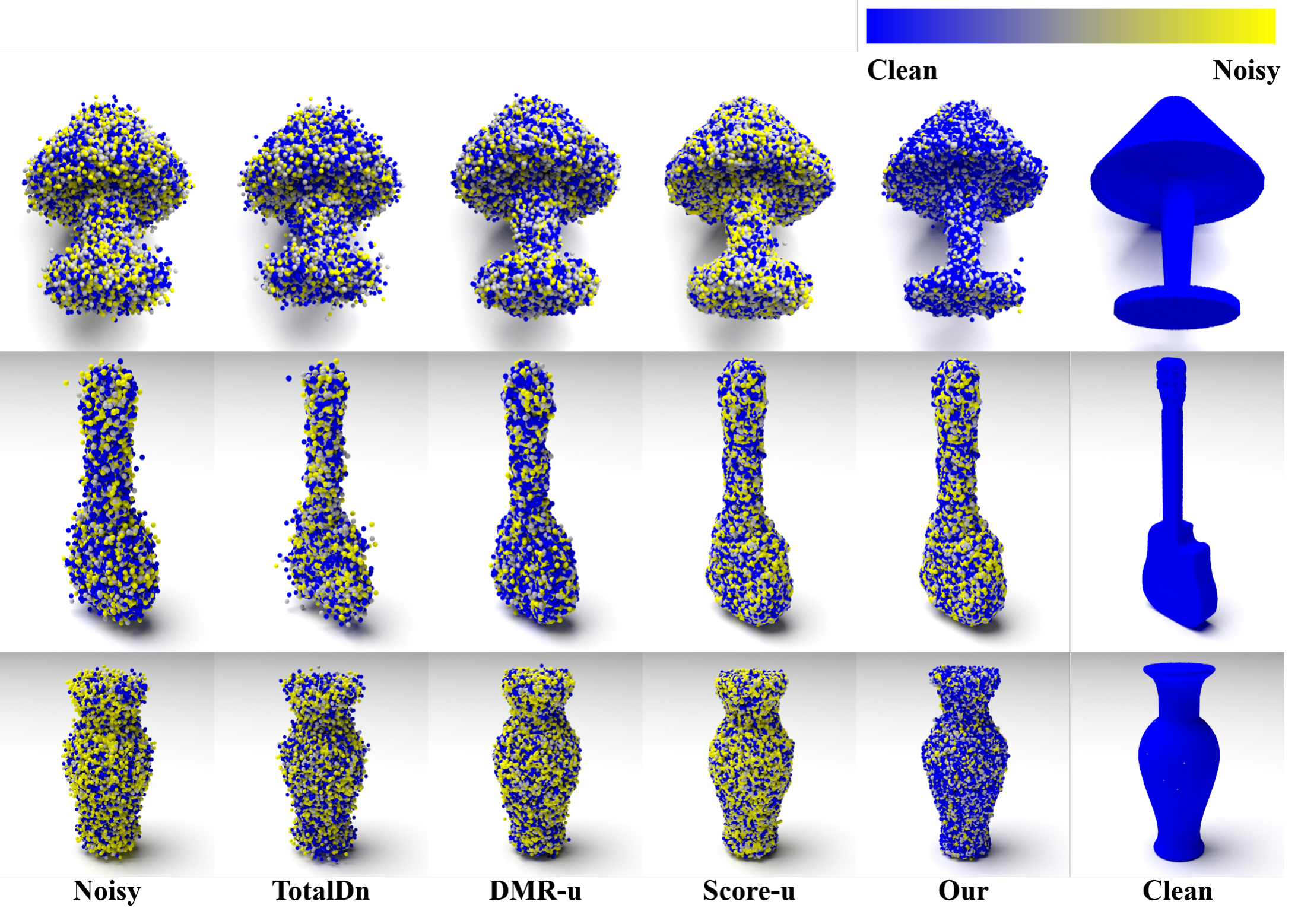}
\end{center}
\vspace{-1.0em}
\caption{Additional visualization results of different algorithms on ModelNet-40 dataset with Gaussian noise. The noise level is set to 3\%.}
\label{fig:suppl_modelnet3}
\end{figure*}

\begin{figure*}[!ht]
\begin{center}
    \includegraphics[width=0.8\textwidth]{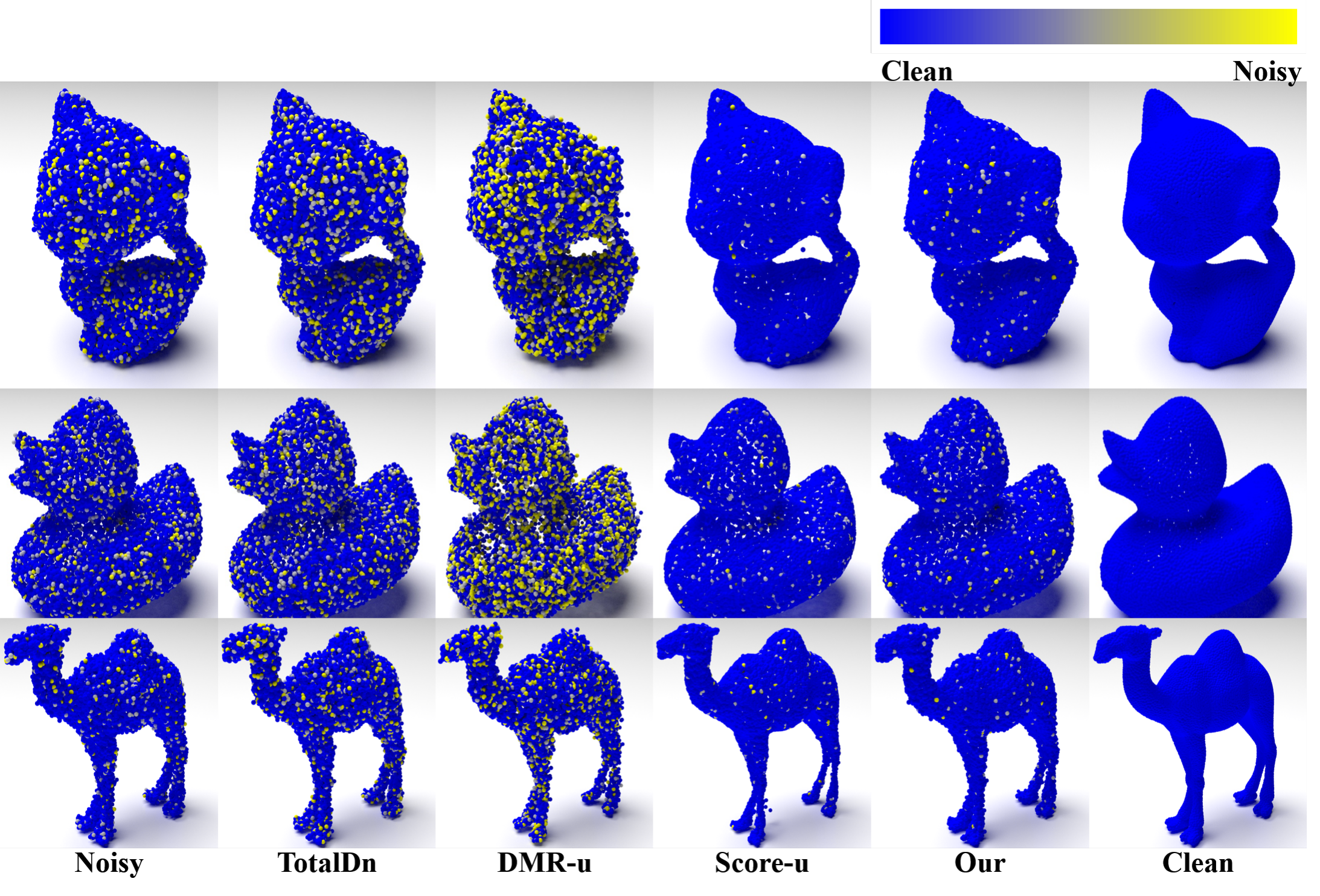}
\end{center}
\vspace{-1.0em}
\caption{Additional visualization results from PU-Net dataset. The noise level is set to 1\%.}
\label{fig:suppl_punet1}
\end{figure*}

\begin{figure*}[!ht]
\begin{center}
    \includegraphics[width=0.8\textwidth]{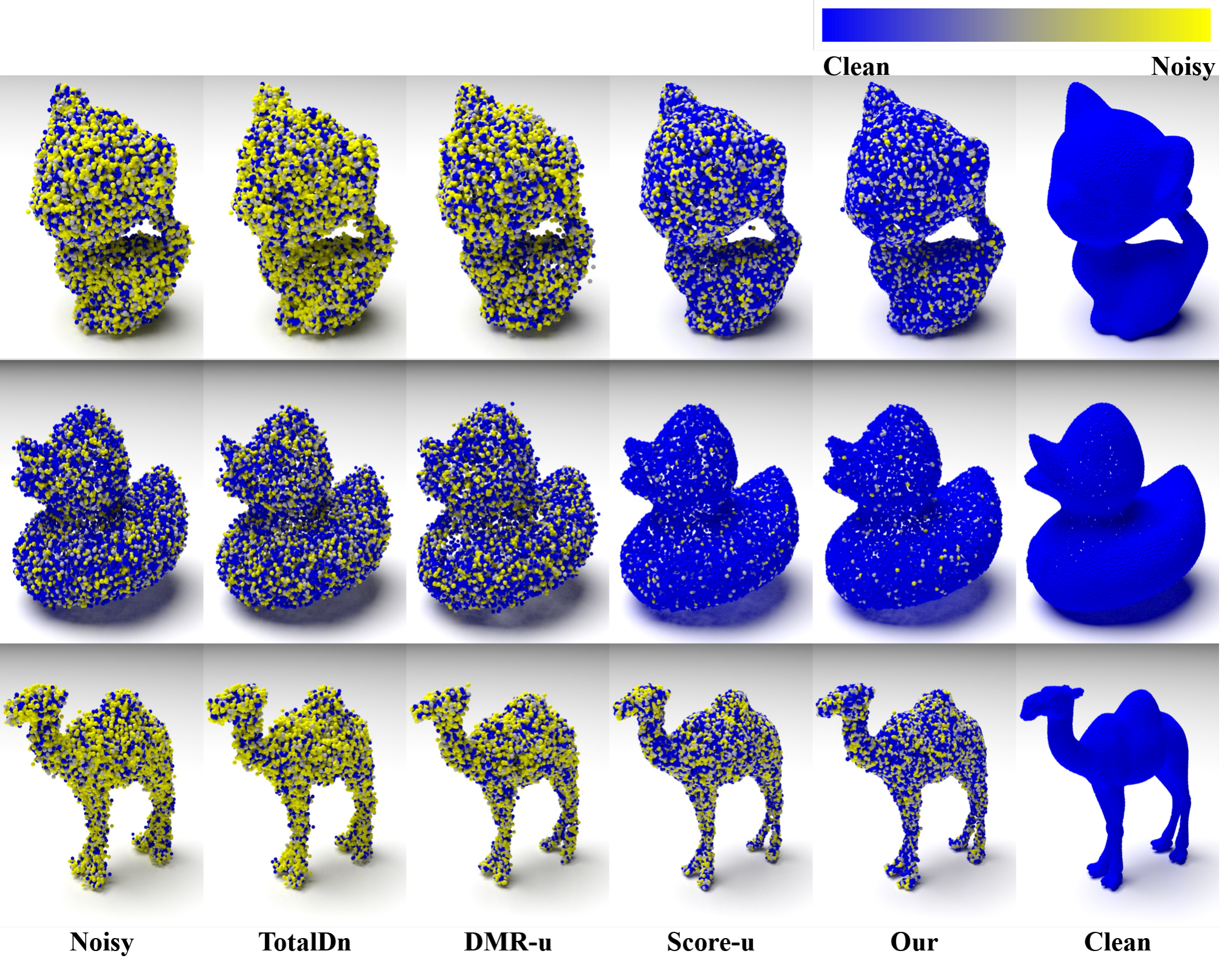}
\end{center}
\vspace{-1.0em}
\caption{Additional visualization results from PU-Net dataset. The noise level is set to 2\%.}
\label{fig:suppl_punet2}
\end{figure*}

\begin{figure*}[!ht]
\begin{center}
    \includegraphics[width=0.8\textwidth]{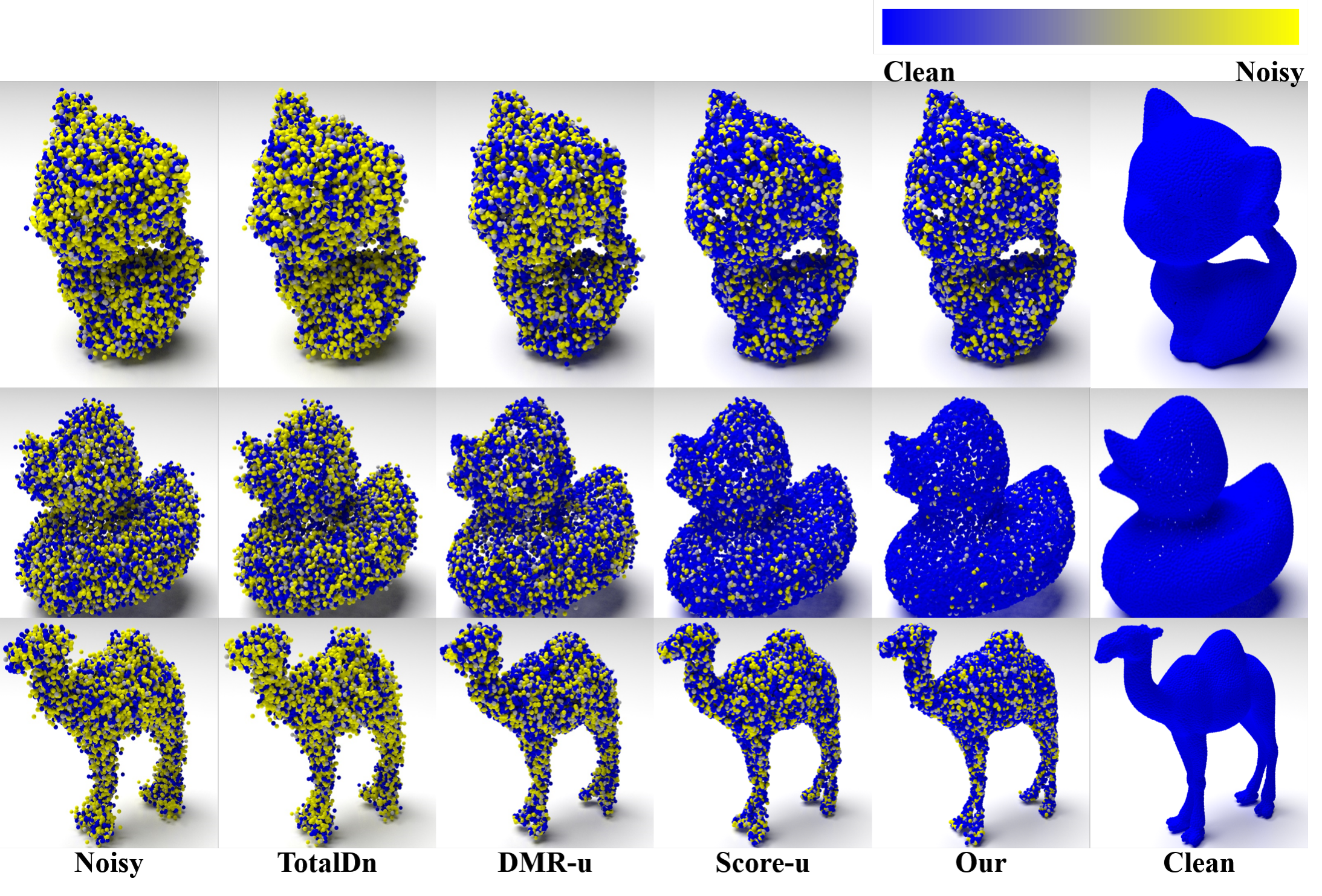}
\end{center}
\vspace{-1.0em}
\caption{Additional visualization results from PU-Net dataset. The noise level is set to 3\%.}
\label{fig:suppl_punet3}
\end{figure*}

\begin{figure*}[!ht]
\begin{center}
    \includegraphics[width=0.8\textwidth]{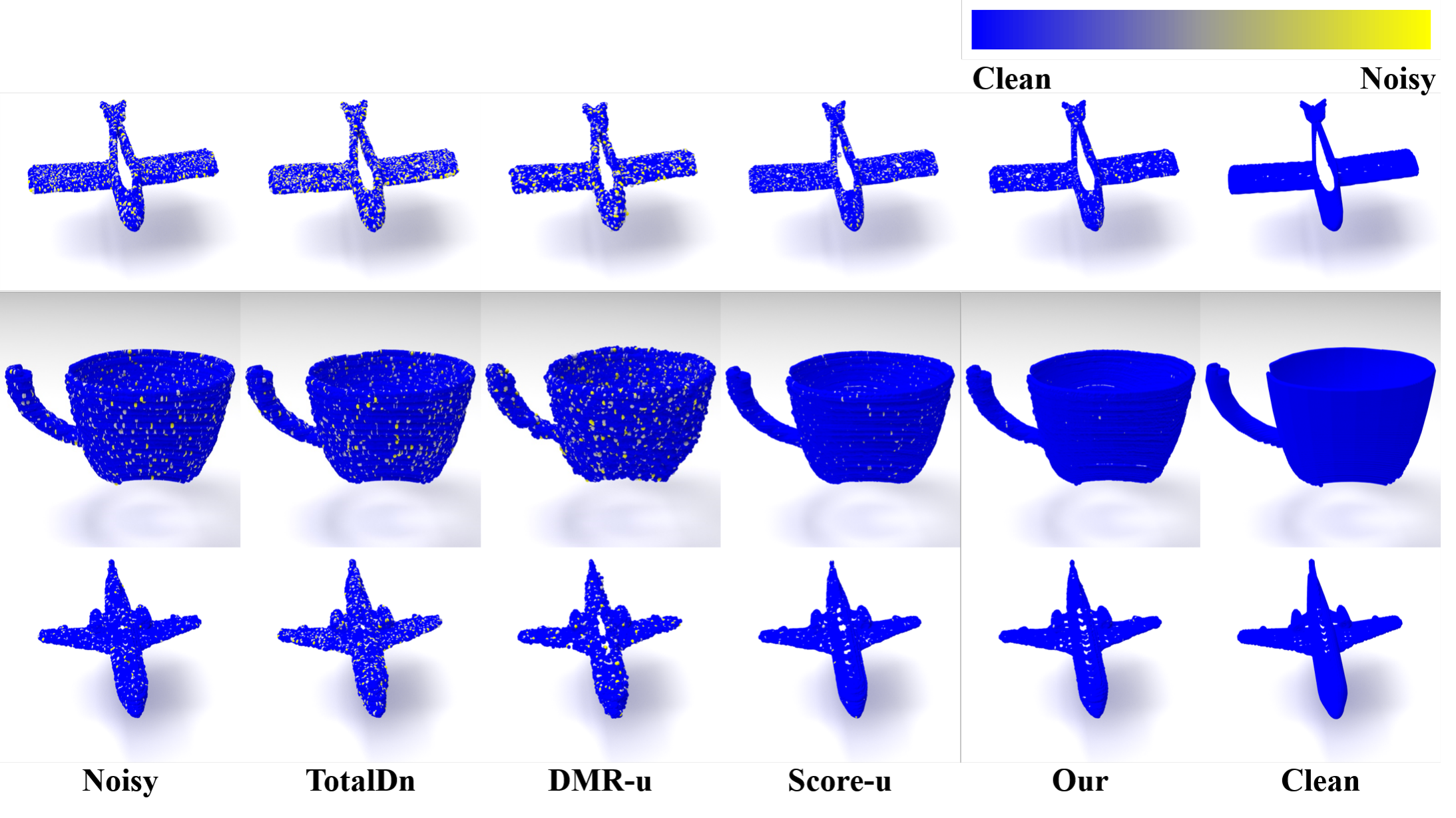}
\end{center}
\vspace{-1.0em}
\caption{Visual comparison of additional denoising results from ModelNet-40 dataset with simulated LiDAR noise. The noise level is set to 0.5\%.}
\label{fig:suppl_modelnetsim05}
\end{figure*}

\begin{figure*}[!ht]
\begin{center}
    \includegraphics[width=0.8\textwidth]{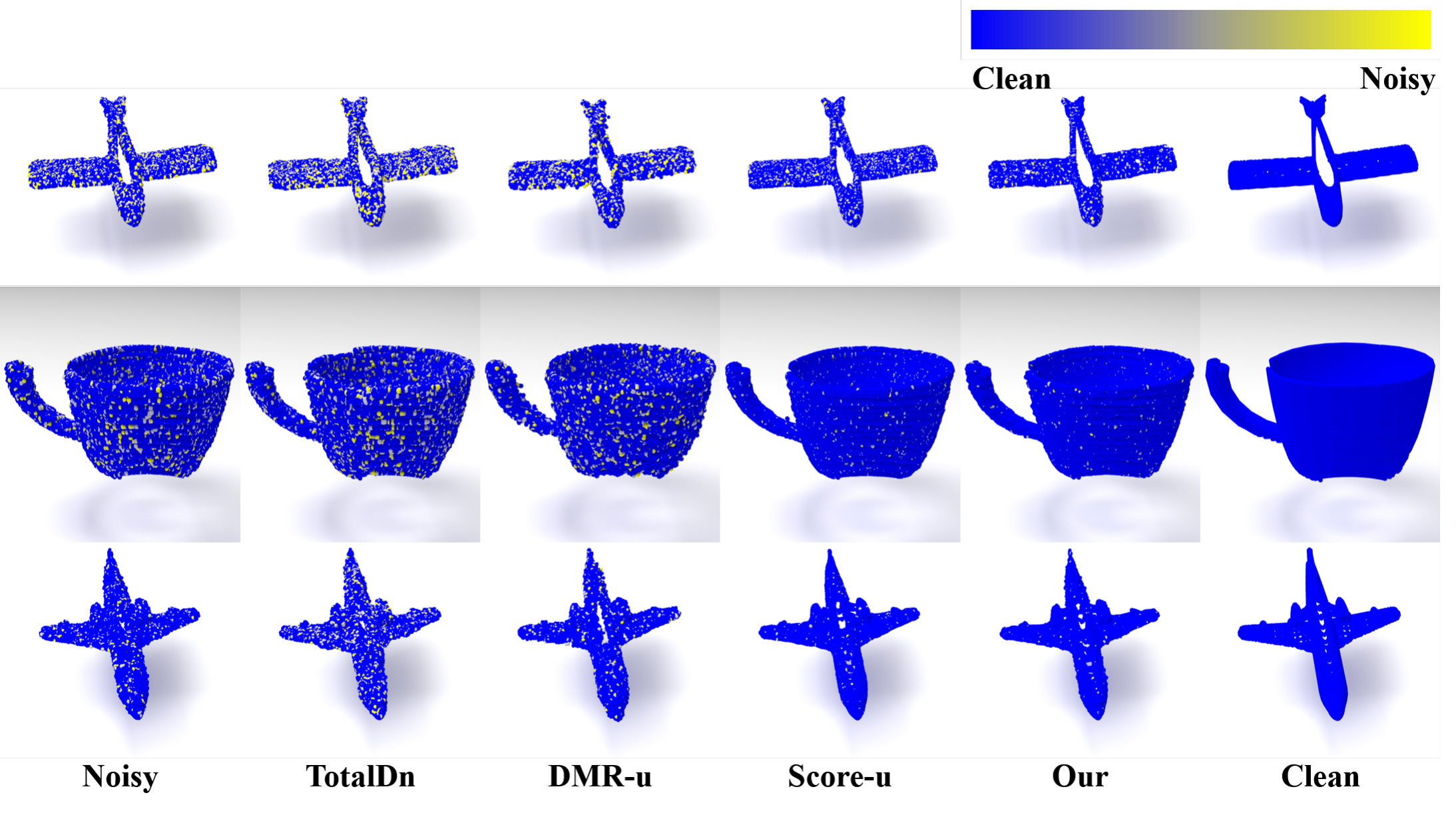}
\end{center}
\vspace{-1.0em}
\caption{Visual comparison of additional denoising results from ModelNet-40 dataset with simulated LiDAR noise. The noise level is set to 1\%.}
\label{fig:suppl_modelnetsim10}
\end{figure*}

\begin{figure*}[!ht]
\begin{center}
    \includegraphics[width=0.8\textwidth]{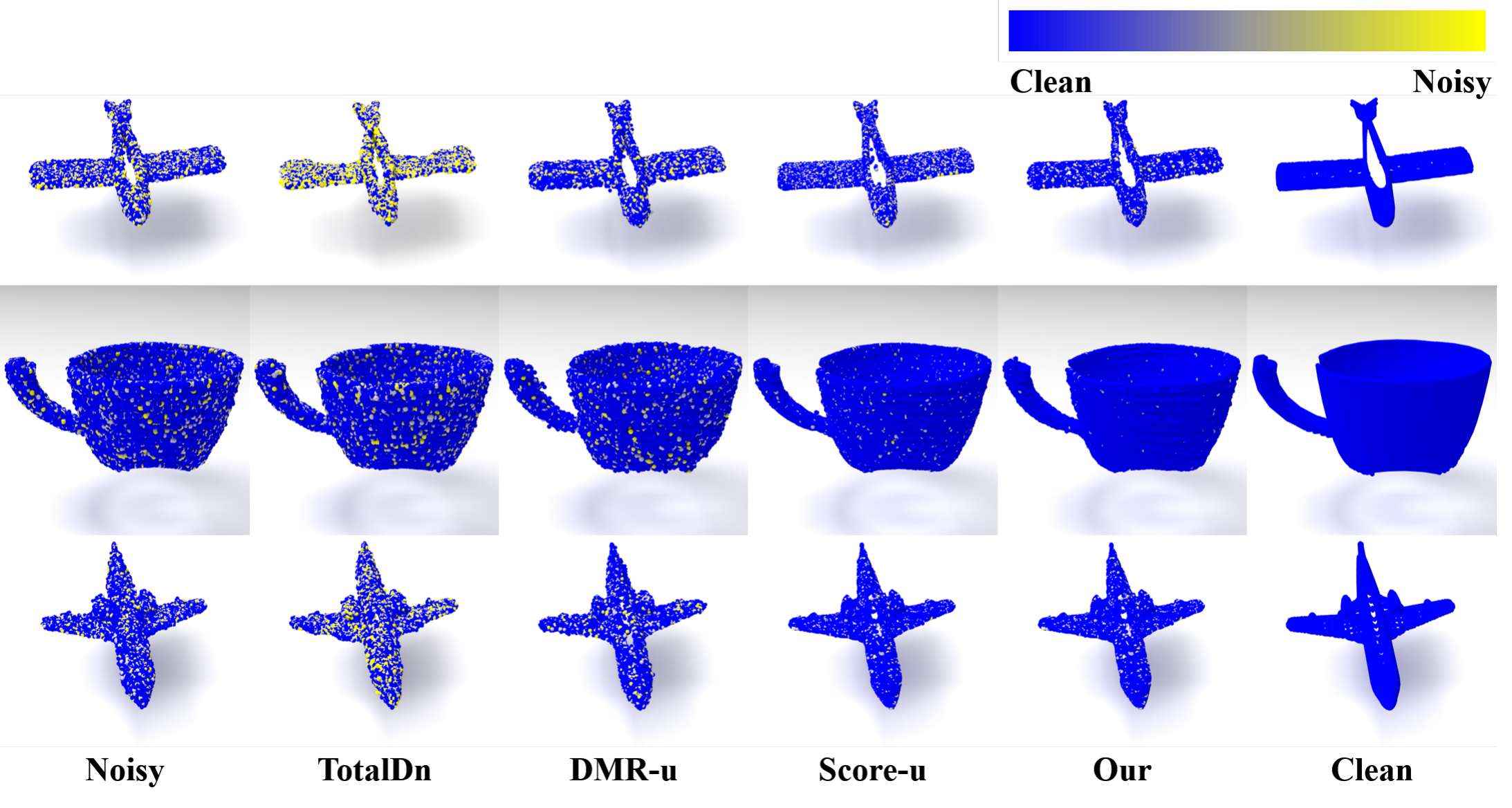}
\end{center}
\vspace{-1.0em}
\caption{Visual comparison of additional denoising results from ModelNet-40 dataset with simulated LiDAR noise. The noise level is set to 1.5\%.}
\label{fig:suppl_modelnetsim15}
\end{figure*}

\cref{fig:suppl_modelnet1,fig:suppl_modelnet2,fig:suppl_modelnet3,fig:suppl_modelnetsim05,fig:suppl_modelnetsim10,fig:suppl_modelnetsim15,fig:suppl_punet1,fig:suppl_punet2,fig:suppl_punet3} provide additional visualization results with varying noise levels on ModelNet-40 and PU-Net. Points with smaller error are colored more blue, and otherwise colored yellow. 
\vspace{2.0em}

%-------------------------------------------------------------------------
\vspace{-1.0em}

\vspace{-1.0em}

\vspace{-1.0em}
%-------------------------------------------------------------------------
%-------------------------------------------------------------------------
\vspace{-1.0em}

\vspace{-1.0em}

\vspace{-1.0em}
%-------------------------------------------------------------------------
%-------------------------------------------------------------------------
\vspace{-1.0em}

\vspace{-1.0em}

%-------------------------------------------------------------------------

\end{document}